\documentclass{article}
\PassOptionsToPackage{numbers}{natbib}
\usepackage[preprint]{neurips_2022}
\usepackage{microtype}
\usepackage{graphicx}
\usepackage{booktabs} 
\usepackage{subfig}

\usepackage{hyperref}

\usepackage{amsmath}
\usepackage{amssymb}
\usepackage{mathtools}
\usepackage{amsthm}

\usepackage[capitalize,noabbrev]{cleveref}

\theoremstyle{plain}
\newtheorem{theorem}{Theorem}[section]

\theoremstyle{definition}
\newtheorem{definition}[theorem]{Definition}

\theoremstyle{remark}

\def\BE{\begin{equation}}
\def\EE{\end{equation}}
\def\BEA{\begin{eqnarray}}
\def\EEA{\end{eqnarray}}

\newenvironment{proofsketch}[1][\proofname]{%
  \proof[Proof Sketch]%
}{\endproof}

\newcommand{\E}{\mathbb{E}}

\newcommand{\abs}[1]{\left|#1\right|}

\newcommand{\inprod}[1]{\left<#1\right>}

\usepackage[textsize=tiny]{todonotes}
\usepackage{graphicx}
\usepackage{placeins}

\title{What do CNNs Learn in the First Layer and Why? A Linear Systems Perspective}
\author{%
  Rhea Chowers \\
  School of Computer Science and Engineering\\
  The Hebrew University\\
  Jerusalem, Israel \\
  \texttt{rhea.chowers@mail.huji.ac.il} \\
   \And
   Yair Weiss \\
  School of Computer Science and Engineering\\
  The Hebrew University\\
  Jerusalem, Israel \\
   \texttt{yair.weiss@mail.huji.ac.il}
}

\begin{document}

\maketitle




\begin{abstract}
It has previously been reported that the representation that is learned in the first layer of deep Convolutional Neural Networks (CNNs) is highly consistent across initializations and architectures. In this work, we quantify this consistency by considering the first layer as a filter bank and measuring its energy distribution. We find that the energy distribution is very different from that of the initial weights and is remarkably consistent across random initializations, datasets, architectures and even when the CNNs are trained with {\em random labels}.  In order to explain this consistency, we derive an analytical formula for the energy profile of linear CNNs and show that this profile is mostly dictated by the second order statistics of image patches in the training set and it will approach a whitening transformation when the number of iterations goes to infinity. Finally, we show that this formula for linear CNNs also gives an excellent fit for the energy profiles learned by commonly used {\em nonlinear} CNNs such as ResNet and VGG, and that the first layer of these CNNs indeed perform approximate whitening of their inputs. 
\end{abstract}

\section{Introduction}

The remarkable success of Convolutional Neural Networks (CNNs) on a wide variety of image recognition tasks is often attributed to the fact that they learn a good representation of images. Support for this view comes from the fact that very different CNNs tend to learn similar representations and that features of CNNs that are trained for one task are often useful in very different tasks \citep{bengio_transfer,predicting_rotations,doimo}. 

A natural starting point for investigating representation learning in deep CNNs is the very first layer. Studying this representation is somewhat easier than studying more general representation learning for the simple reason that the output of this layer is a linear function of its input. Thus we can use the perspective of linear systems whereby a system based on convolutions can be fully characterized by its frequency response. In this paper, we adopt the linear systems perspective and consider the first layer as a filter bank and measure the sensitivity of the bank to different spatial frequencies. As we show in section~\ref{section:consistency}, this profile of sensitivities (which we call the "energy profile") is highly consistent for different initializations,  architectures and training sets and is very different from the profile of the initial random weights. The filter bank's sensitivity peaks at intermediate spatial frequencies, while being \textit{insensitive} to very high and low spatial frequencies.

The linear systems perspective has been used in the past to analyze biological neural networks \citep{AtickRedlich90} where it has been argued based on first principles that the first layer of a  neural network should perform  "redundancy reduction" \citep{barlow}. For example, in the case of images, the pixel representation is highly redundant since neighboring pixel values are highly correlated. Under the redundancy reduction hypothesis, the goal of early layers is to "disentangle" the input and remove these correlations to facilitate downstream learning.  When this hypothesis is formalized,  the resulting optimal transformation takes the form of "whitening": the sensitivity of the first layer to a particular frequency should be inversely proportional to the variance of the input signal at that frequency (provided that the input variance is much larger than the noise).  Such "whitening" transformations have been observed experimentally in different biological systems \citep{natim}, and several authors have recently argued that whitening should be enforced in the different layers of CNNs \citep{zca_batch_normalization,Zhang_2021_CVPR}.

 \begin{figure*}[h]
  \centering
  \subfloat[ CIFAR10] {\label{fig:cifar10norandom}\includegraphics[width=0.45\linewidth]{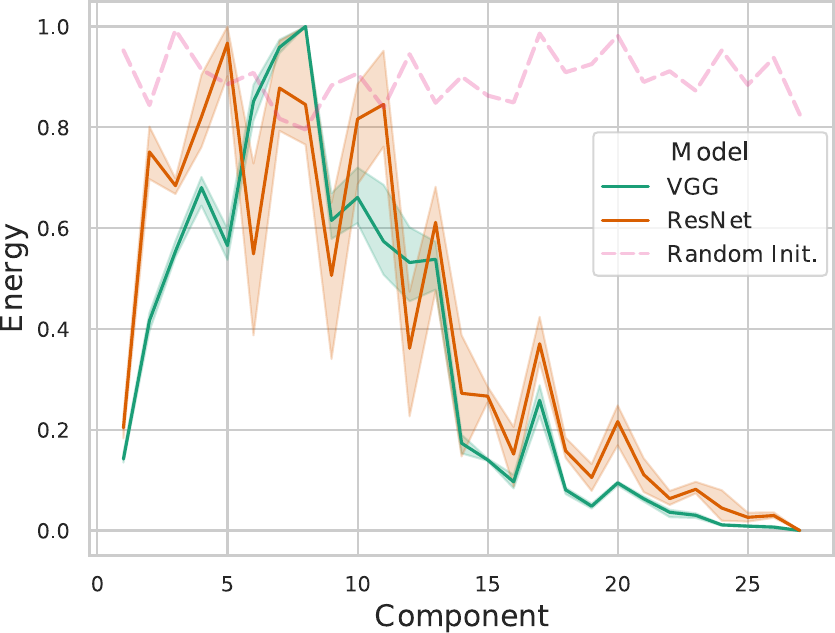} }%
  \hfill
    \subfloat[ CelebA]{{\label{fig:celebanorandom}\includegraphics[width=0.45\linewidth]{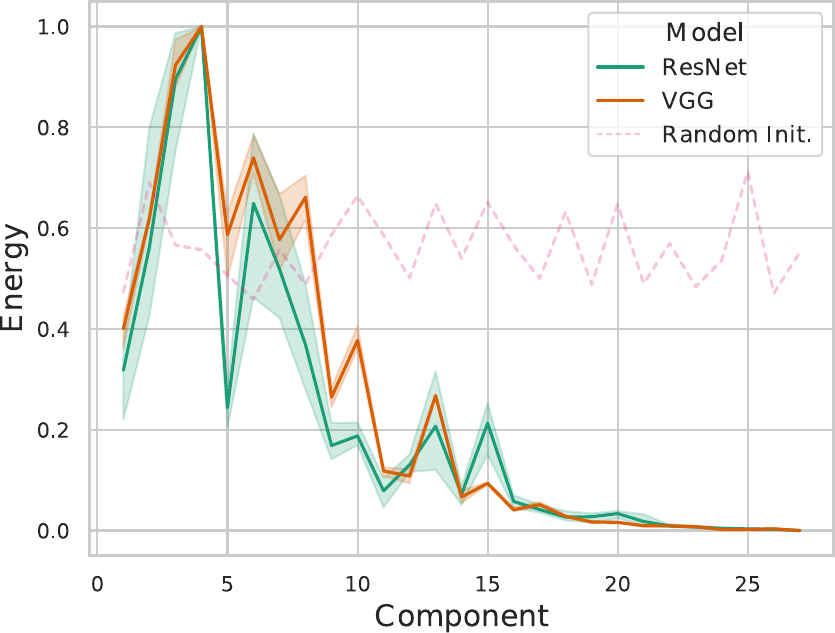} }}%
  \hfill
  \subfloat[ Imagenet]{\label{fig:}\includegraphics[width=0.45\linewidth]{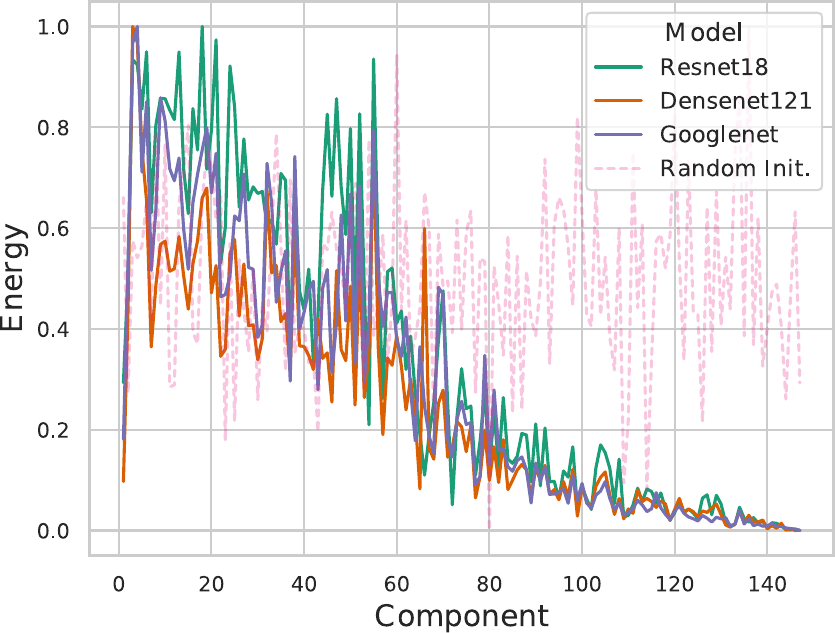} }%

  \caption{When trained on various datasets, different models both learn consistent energy profiles. CIFAR10 and CelebA are averaged over many different initializations and the spread indicates the variance. Models trained on ImageNet were downloaded from the Pytorch library. See \cref{table:comparisons} for full correlation coefficients. An example of a random initialization is plotted for reference.}
  \label{fig:consistency}%
   \vskip 0.1in
\end{figure*}

If CNNs were trained with an explicit "redundancy reduction" loss function, we would therefore expect their energy profiles to be consistent for different architectures and random initializations, but why does this consistency occur when the networks are trained to minimize a classification loss on the training set? A possible explanation  is that these filters are optimal in some sense for solving the recognition task.  Thus, the networks have simply learned that in order to minimize the training loss the first layer of deep CNNs must have filters whose energy profile has a particular shape.

In this paper we present empirical and theoretical results that are \textbf{inconsistent with this explanation}. We show that trained networks learn consistent representations that are far from their initialization despite the fact that  CNNs with commonly used architectures can be trained with \textbf{fixed, random filters in the first layer} and still yield comparable performance to full learning. We also show that the same energy profile is obtained when the network is trained to predict \textbf{random labels}. We then show that under realistic assumptions on the statistics of the input and labels, consistency also occurs in simple, linear CNNs, and derive an analytical form for its energy profile. We show that as the number of iterations goes to infinity,  this profile takes the form of a first layer that performs whitening of the input image patches. Finally, we show that the analytical formula which we derived for linear CNNs gives an excellent fit to the energy profile \textbf{of real-world CNNs} as well, when trained with either \textbf{true or random labels}.

\section{Quantifying Consistency using Energy}\label{section:consistency}

Defining the similarity between the representations learned by different CNNs is challenging~\citep{oldsimilarity, subspacematching}. The dimension of the representation  may be different and even when they are the same, the two representations may be very different when individual neurons are compared but still identical when the full representation is compared (e.g. two representations that are rotations of each other). 
Recent works \citep{CKA, minibatchCKA} suggest comparing two representations based on the distance between the distribution over patches induced by the two representations. But estimating this distance in high dimensions is nontrivial and two very different networks might give similar distributions over patches when the input distribution is highly skewed~\cite{ding2021grounding}. In this paper we propose a new method which avoids these shortcomings and is especially relevant for the first layer of a CNN. 

Our method is based on the linear systems perspective, whereby a system that is based on convolutions is fully specified by its frequency response. Since the filters in CNNs are typically highly localized in space (e.g. many successful CNNs use $3 \times 3 \times 3$ filters in the first layer) we characterize this frequency response using the principal components of the input image patches. 


\begin{definition}
Given a set of patches $\{p_n\}$ the PCA vectors $u_i$ are eigenvectors of the matrix $\sum_{n} p_n p_n^T$. 
\end{definition}
\begin{definition}Given a set of filters $\{w_k\}$ and a set of PCA
vectors $\{u_i\}$ the \textbf{energy profile}
of the set is given by a vector $e$ whose $i$th component is given by:
\BE
e_i^2 = \frac{1}{K} \sum_{k=1}^K (w_k^T u_i)^2
\EE
\end{definition}



We measure similarity between two different sets of filters by measuring the correlation coefficient between their energy profiles. Note that this measure is invariant to a rescaling of the filters, to a permutation of the filters and to any orthogonal transformation of the filters. Since the PCA vectors of a set of patches extracted from natural images are highly localized in frequency~\cite{natim}, 
this way of comparing linear representation is equivalent to considering the set of filters as a filter bank and measuring the sensitivity of the filter bank to different spatial frequencies. 

\begin{table}[h]
\caption{Correlation between energy profiles of VGG11, trained with different random seeds (initializations), first layer widths, over various datasets, and compared with ResNet18. Standard deviation provided in cases it exceeds $0.04$.}
\label{table:comparisons}
\vskip 0.15in
\begin{center}
\begin{small}
\begin{sc}



\begin{tabular}{lccccr}
\toprule
Dataset & Seed & Width & Trained vs  & Vgg vs \\
 &  & & Init  & Resnet \\

\midrule

Cifar10&0.99&0.98& -0.13 $\pm$ 0.18&0.87 \\
Cifar100&0.97&0.98& -0.04 $\pm$ 0.04&0.80 \\
CelebA&0.99 &0.98& -0.18 $\pm$ 0.13&0.92  \\

\bottomrule
\end{tabular}
\end{sc}
\end{small}
\end{center}
\vskip 0.1in
\end{table}

\cref{fig:consistency} show that different models trained with gradient descent are remarkably consistent using our proposed measure. Regardless of architecture or the particular dataset that they were trained on, different CNNs have very similar energy profiles that are less sensitive to very high or low spatial frequencies, and the peak sensitivity is for intermediate spatial frequencies. This profile is very different from the profile of the initial, random, filters which is approximately constant for all frequencies. 
\cref{table:comparisons} quantifies this similarity. The correlation coefficient between energy profiles of trained models with different random initializations and architecture is remarkably high (over 0.98 in the case of different seeds and first layer widths) and the correlation between the learned profiles and the random initialization is close to zero. An extensive set of experiments on various models and datasets can be found in \cref{appendix:pretrainedcors}.

 Thus the use of our new measure allows us to quantitatively show that deep CNNs trained with gradient descent using standard parameters exhibit highly consistent representation, namely in the form of sensitivity to intermediate spatial frequencies. We now ask: what determines this consistency?

\section{Is Consistency due to CNNs Learning Semantically Meaningful Features?}\label{section:meaningfulfeatures}

A natural explanation for the remarkable consistency of the learned representation in the first layer is that CNNs learn a representation that is good for object recognition. In particular, high spatial frequencies are often noisy while very low spatial frequencies are often influenced by illumination conditions. Thus learning a representation that is mostly sensitive to intermediate spatial frequencies makes sense if the goal is to recognize objects. Similarly, human vision is also mostly sensitive to intermediate spatial frequencies \citep{CSF}, presumably for the same reasons. 

In order to test this hypothesis we asked if training modern CNNs while freezing the first layer will result in a decrease in performance. If indeed a set of filters that is sensitive mostly to intermediate frequencies is optimal for object recognition, we would expect performance to suffer if we froze the first layer to have random filters with equal energy in all frequencies. 
\begin{figure}[h]
   \centering
  \subfloat[\centering Training Loss]{{\includegraphics[width=0.45\linewidth]{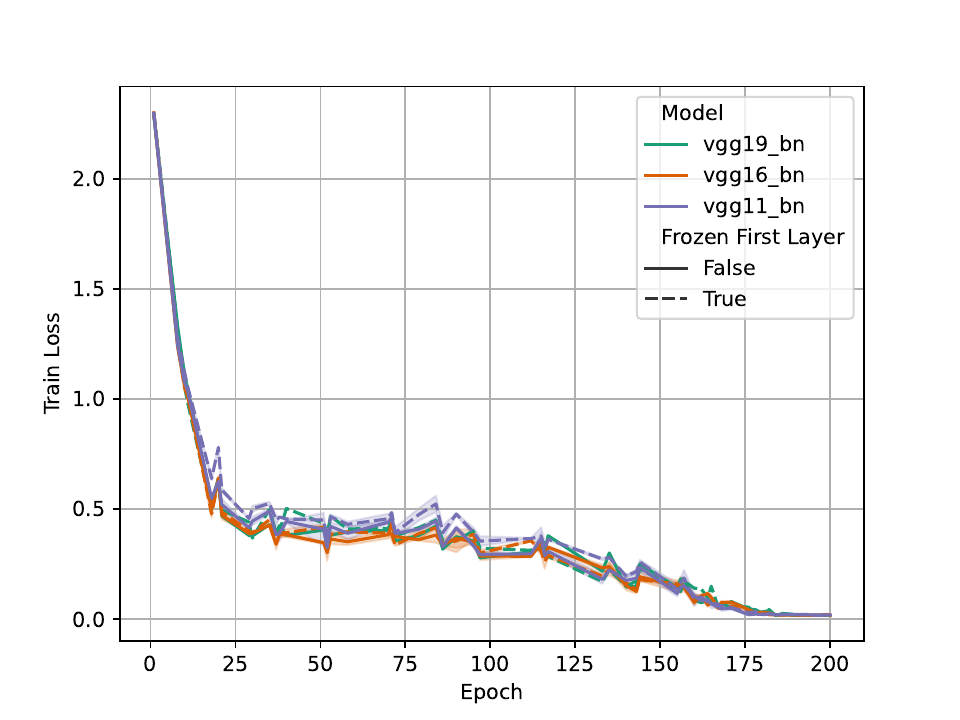} }}%
  \hfill
  \subfloat[\centering Validation Loss]{{\includegraphics[width=0.45\linewidth]{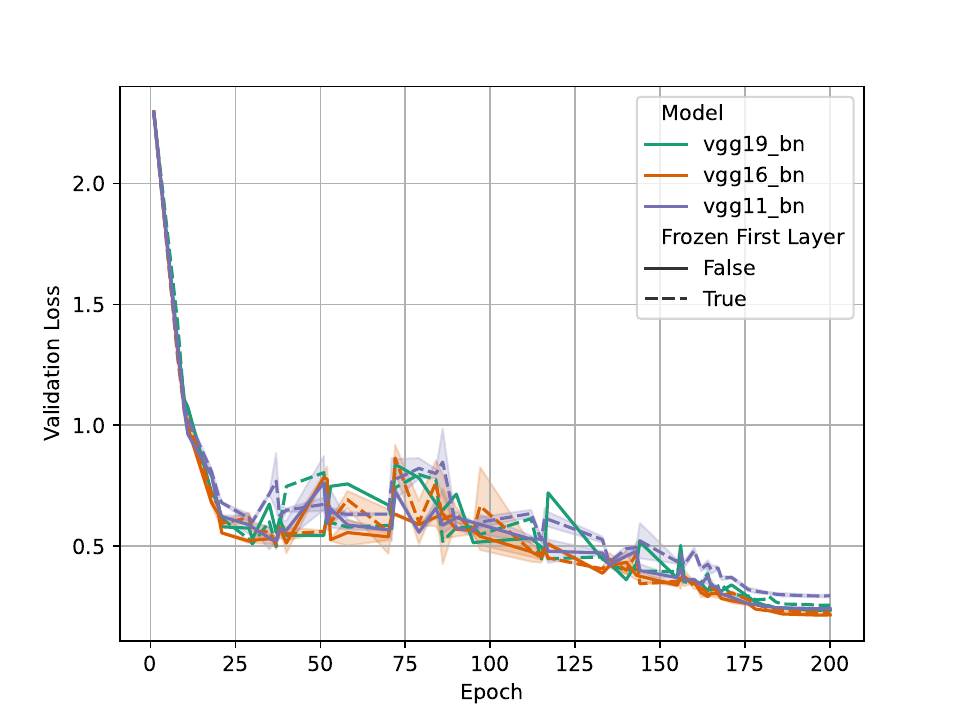} }}%
  
   \caption{Training and validation loss of VGGs of different depths on CIFAR10 as function of iteration with frozen first layer and without. For deep networks the performance is the same as with frozen layer. Accuracy figures can be found in \cref{frozen_appendix}}\label{freeze}
   \vskip 0.1in
\end{figure}
\begin{figure*}[t]
  \centering
    
  \subfloat[\centering CIFAR10]{{\label{fig:cifar10random}\includegraphics[width=0.45\linewidth]{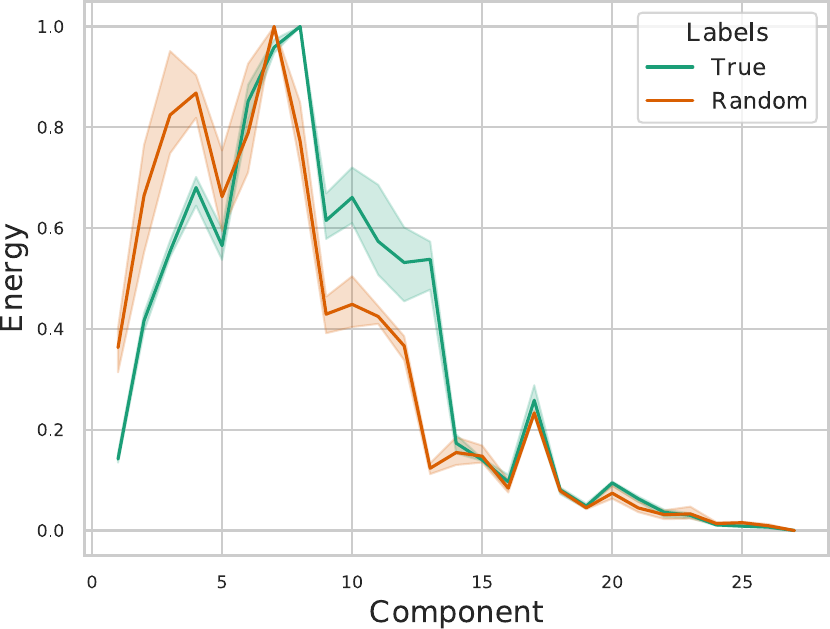} }}%
  \hfill
  \subfloat[\centering CIFAR100]{{\label{fig:cifar100random}\includegraphics[width=0.45\linewidth]{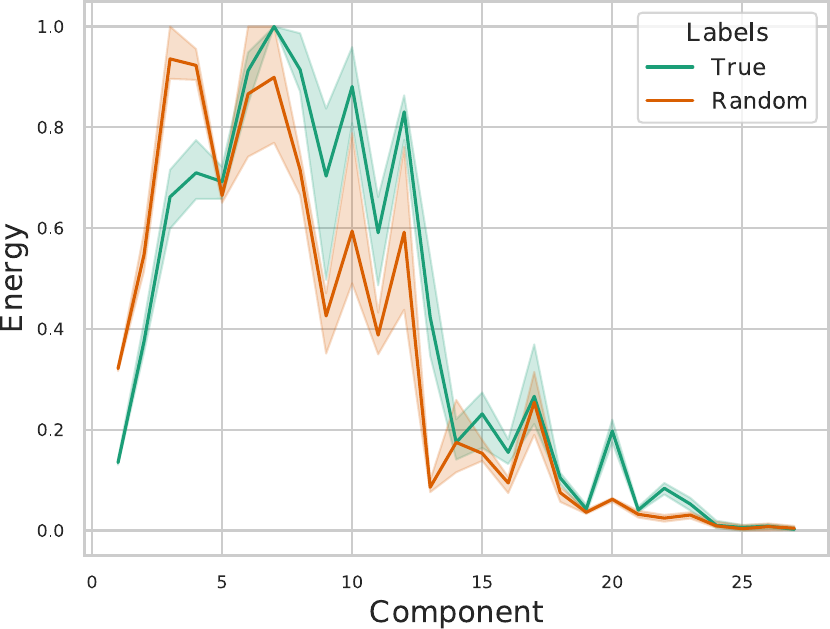} }}%
  \hfill
  \subfloat[\centering CelebA]{{\label{fig:facesrandom}\includegraphics[width=0.45\linewidth]{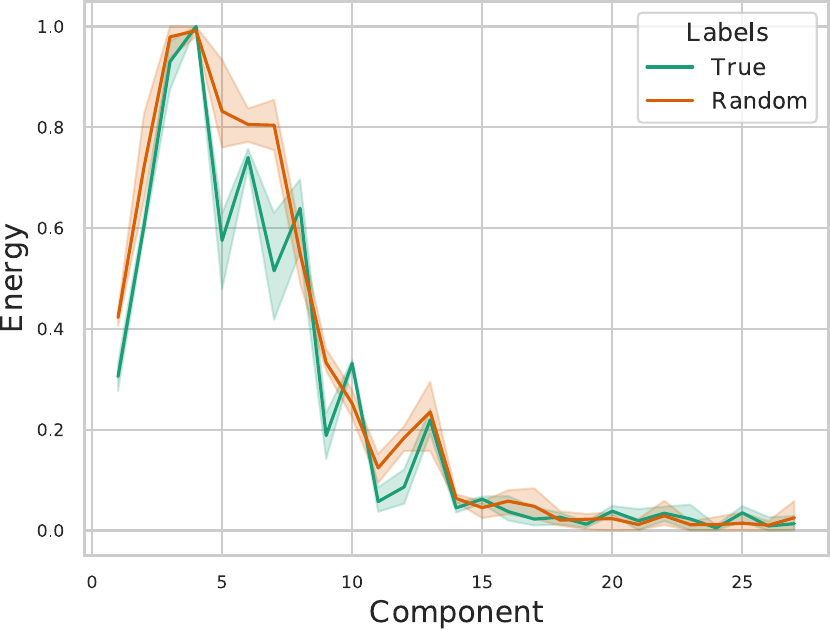} }}%
  \caption{VGG11 trained on CIFAR10 (\cref{fig:cifar10random}) CIFAR100 (\cref{fig:cifar100random}) and a CelebA classification task (\cref{fig:facesrandom}) exhibit similar energy patterns when trained with true and random labels. These are also highly correlated and differ from initialization (see \cref{table:truerandom}). Further experiments on binary CIFAR10 subsets and with ResNet can be found in \cref{appendix:pretrainedcors}.}%
  \label{fig:vggtruerandom}%
\end{figure*}

\cref{freeze} shows that there is almost no change in the performance of modern CNNs when the weights in the first layer are frozen. This is true when measuring training accuracy, training loss or validation accuracy and validation loss (and see \cref{frozen_appendix}). Apparently the networks learn to compensate for the random filters in the first layer by learning different weights in the subsequent layers. In other words, if we were to train modern CNNs using some discrete search over weights (e.g. genetic programming) to minimize the training loss, there is no reason to expect a consistent energy profile that is sensitive mostly to intermediate spatial frequencies to be found.  Equally good training loss can be obtained with random filters in the first layer.

Another test to this hypothesis can be done by training networks with random labels. In this setting, models are known to memorize their training set \citep{memorization}. While a particular energy profile may be optimal for recognizing natural object categories (e.g. for ignoring illumination effects), we should not expect any particular set of features to be optimal for recognizing randomly defined categories. Surprisingly, however, we find {\bf the same energy profile when CNNs are trained with true labels and random labels.} \cref{fig:vggtruerandom} compares the energy profiles of models trained with true and random labels on different datasets, and shows a highly consistent profile between the two sets of labels. \cref{table:truerandom} shows that this result is consistent over multiple random seeds and far from initialization. 

To summarize, while quantitatively highly consistent representations are learned in the first layer of commonly used CNNs, this cannot be explained by the networks minimization of the training loss. Furthermore, the learned set of features is consistent for models trained with \textbf{random labels} as well, suggesting a bias in the input, training algorithm, or both. This motivates us to analyze representation learning in much simpler CNNs.

\begin{table}[h]
\caption{Correlation between energy profiles of VGG11 trained with true and random labels for different datasets. While highly correlated, the profiles are far from initialization. }
\label{table:truerandom}
\vskip 0.15in
\begin{center}
\begin{small}
\begin{sc}
\begin{tabular}{lcccr}
\toprule
Dataset &  VGG (Random) &  VGG (True) vs \\
 &  v Init &  VGG (Random) \\

\midrule

Cifar10&0.03 $\pm$ 0.22&\textbf{0.90 $\pm$ 0.02}\\
Cifar100&0.14 $\pm$ 0.13&\textbf{0.91 $\pm$ 0.01}\\
CelebA&0.08 $\pm$ 0.07&\textbf{0.96 $\pm$ 0.03}\\
\bottomrule
\end{tabular}
\end{sc}
\end{small}
\end{center}
\vskip 0.1in
\end{table}

\section{Theory in Simple Linear CNNs}\label{theoretical_section}
In order to understand the consistency that we observe among energy profiles in the first layer of trained CNNs, we turn to analyzing a very simple model: a {\em linear} CNN with one hidden layer trained with the MSE loss. Specifically, in this simple model, the first layer includes convolutions with $K$ different filters and the output is given by a global average pool of the filters over all locations. 

This model is clearly very different from real-world CNNs, but it allows a closed form analysis of the energy profile in the first layer. Furthermore, we will subsequently show that it exhibits many of the same properties as those of real-world CNNs. 

Our main theorem (\ref{theorem:maintheorem}) provides an analytic formula for the energy profile of these models, which is consistent across initializations and widths of the first layer. Additionally, given that true labels are uncorrelated with image patches, the theorem implies consistency between models trained with true and random labels as well. 

The theorem relates the energy profile of the learned filters to the energy profile of the training patches, which we now define.

\begin{definition}Given a set of patches $\{p_n\}$ and a set of PCA
vectors $\{u_i\}$ the energy profile
of the set is given by a vector $\lambda$ whose $i$th component is given by:
\BE
\lambda_i^2 = \frac{1}{N} \sum_{n=1}^N (p_n^T u_i)^2
\EE
\end{definition}

\begin{theorem}\label{theorem:maintheorem}
    Consider a depth-2 linear CNN of any width
initialized with zero mean filters and variance $\sigma^2 I$ and trained
with gradient descent with step size $\eta$ on the MSE loss. Assume that different patches
in each image are uncorrelated with each other and that the labels are
uncorrelated with individual PCA components, then as the number of
patches in the training set goes to infinity, the energy profile of
the filters at iterations $t$ is given by: 

\begin{equation}
  \label{to-provew-eq-main}
  e_i  = \tilde{c}\cdot \frac{|1-(1-\eta \lambda_i^2)^t|}
    {\eta^2 \lambda_i^2} \lambda_i + \xi_{i}
    \end{equation}
where $\lambda_i$ is the energy profile of the training patches and
$\xi$ a random vector that depends on the initialization and whose
magnitude goes to zero as $\sigma \rightarrow 0$ .
\end{theorem} 
\begin{proofsketch}
The result is obtained by explicitly calculating the gradient of the  MSE loss with respect to the average filter and noting that the dynamics of gradient descent can be written as scalar dynamics in PCA space and take the form of a geometric series. The result also uses the assumption that the labels are uncorrelated with the PCA coefficients to obtain a formula that does not depend on the labels. Even though the labels are uncorrelated with the PCA coefficients, any finite dataset will include small, spurious correlations and the magnitude of these correlations will be almost surely proportional to $\lambda_i$.   A full proof is supplied in \cref{appendix:proof}.
\end{proofsketch}
\begin{figure}[h!]
  \centering
   \centerline{\includegraphics[width=0.45\columnwidth]{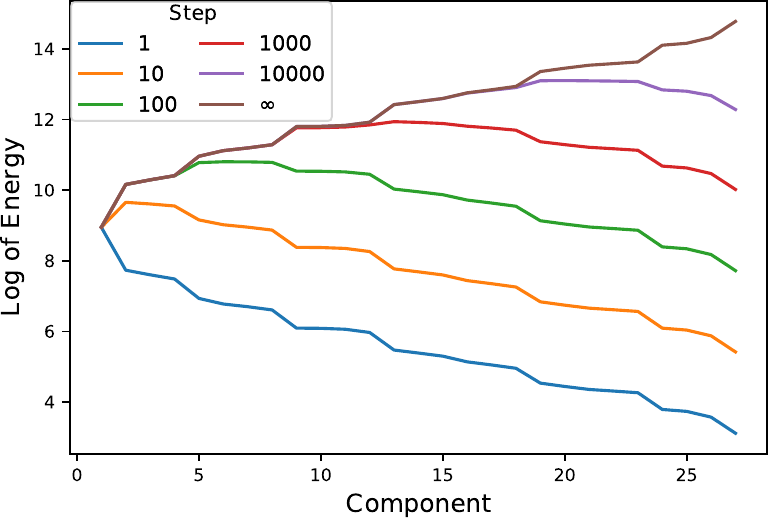}}

  \caption{Analytic formula at different iterations with a constant learning rate, for CIFAR10. At early training iterations the formula is sensitive to the largest eigenvalue (corresponding to the first PCA component). The sensitivity shifts as the number of iterations increases and at each iteration the formula performs whitening on increasingly higher frequencies. }\label{fig:formulasim}
   \vskip 0.1in
\end{figure}

Thus under our assumptions, the energy profile will only depend on the second-order statistics of the input patches (as described by the energy profile $\lambda_i$), the number of iterations, and the learning rate. 
But what does this profile mean?  \cref{fig:formulasim} shows the analytical formula at different iterations with a constant learning rate when the energy profile of the patches $\lambda_i$ is calculated on CIFAR10 (note the log scale on the y axis).  At early training iterations, the formula is mostly sensitive to low spatial frequencies but the sensitivity shifts as the number of iterations increases. As the number of iterations approaches infinity, the profile is actually sensitive mostly to high spatial frequencies. The following theorem shows that as the number of iterations goes to infinity, the filters of a linear CNN perform whitening.

\begin{theorem}\label{theorem:decorrelation}
 Let $\{w_k\}$ be the filters in the first layer of a
CNN. If the energy profile of these filters satisfy:
\begin{equation}
  \label{to-provew-eq-main2}
  e_i  = \tilde{c}\cdot \frac{|1-(1-\eta \lambda_i^2)^t|}
    {\eta^2 \lambda_i^2} \lambda_i 
    \end{equation}
    then as the number of iterations goes to infinity, the
filters in the first layer of the CNN perform spatial decorrelation: the vector of responses at any given location is uncorrelated with the vector of responses at any other location.
\end{theorem} 

\begin{proofsketch}
For any learning rate $\eta<\frac{1}{\max_i{\lambda^2_i}}$, at the limit $t\rightarrow \infty$ then $(1-\eta \lambda_i^2)^t \rightarrow 0$, meaning $e_i\propto \frac{1}{\lambda_i}$, which is a whitening filter and therefore performs spatial decorrelation. For full proof see \cref{appendix:proof}.
\end{proofsketch}

In other words, when assuming that the labels and input patches are uncorrelated - an assumption obviously valid for random labels but requires justification for true ones, simple linear CNNs learn consistent energy profiles which converge to whitening, or performing spatial decorrelation. The profile depends only on the second order statistics of patches, and is consistent for different initializations and widths, and will be the same with true and random labels. For finite iterations, the filters will not perform full whitening and only those components for which $\lambda_i$ is large will be whitened (\cref{fig:formulasim}). This is similar to the optimal redundancy reduction that was derived from first principles in~\cite{AtickRedlich90} and suggested that only components for which $\lambda_i$ is much greater than the noise should be whitened. But unlike the explicit "redundancy reduction" discussed in previous works, here partial whitening emerges due to a trade off with the number of iterations, caused by the use of gradient descent to minimize the training loss.

\begin{figure}[h]
  \centering
    
  \subfloat[\centering CIFAR10]{{\label{cifarlineplot}\includegraphics[width=0.45\columnwidth]{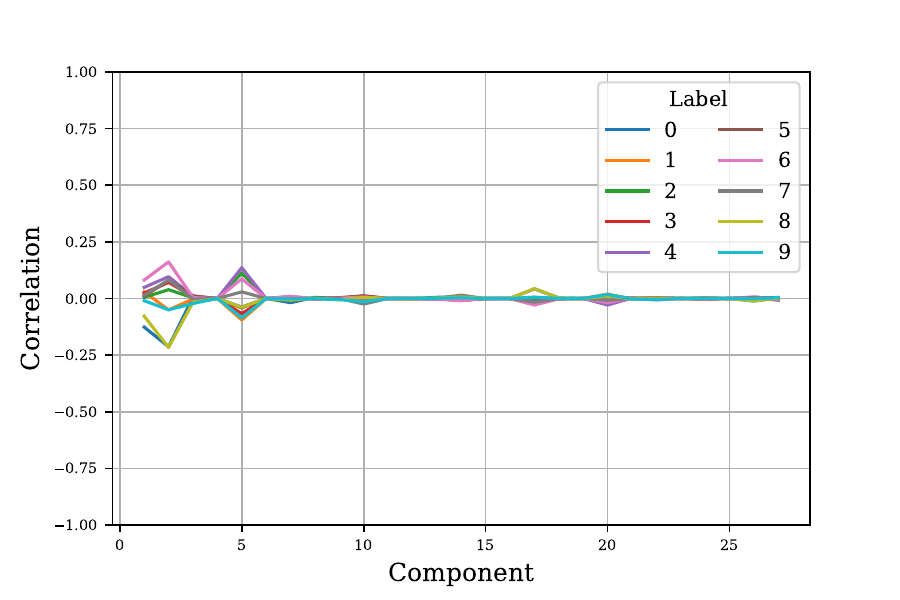} }}%
  \hfill
    \subfloat[\centering ImageNet (10)]{{\label{imagenetlineplot}\includegraphics[width=0.45\columnwidth]{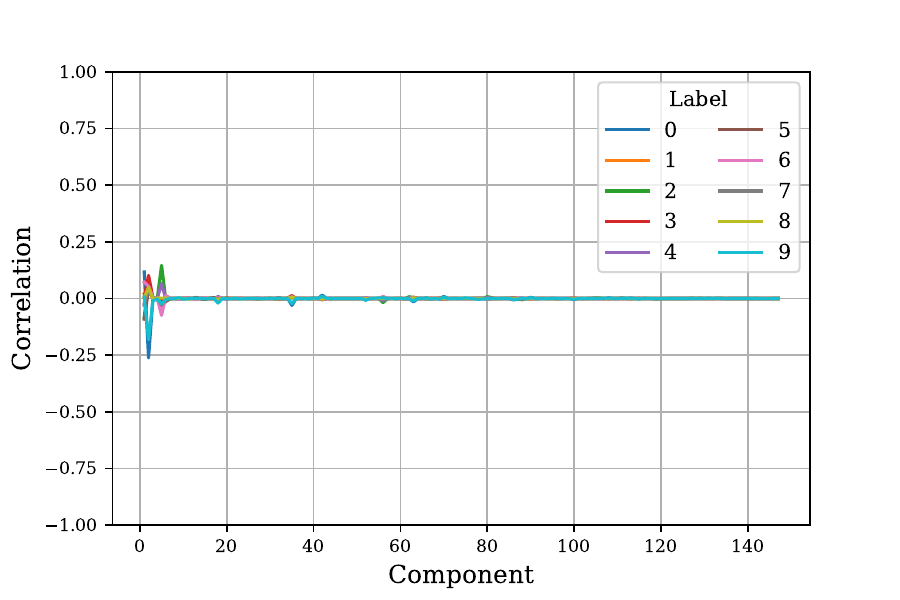} }}
  \caption{Correlation between the patch energy in each PCA component and the class labels for CIFAR10 (\cref{cifarlineplot}, using $3\times 3\times 3$ patches) and a 10 class subset of ImageNet (\cref{imagenetlineplot}, using $3\times 7\times 7$ patches). The label vector is 1 for a given class and zero for all other classes.  Correlations are all around 0, suggesting the assumption that patches are uncorrelated with their labels is true for real datasets.}%
  \label{fig:pcalabels}%
  
\end{figure}

\section{Comparing Theory to Practice}\label{section:formulafit}
The theory in the previous section used a highly simplified CNN trained with MSE loss. We now ask: how well does the theory predict the energy profiles of real-world, nonlinear CNNs trained with the standard cross-entropy loss? 
We first measure the validity of the assumption that the labels are uncorrelated with individual PCA coefficients.  \cref{fig:pcalabels} measures the correlation between the labels (one vs. all) and  PCA coefficients in CIFAR10 and a subset of ImageNet with 10 classes. For all  classes, correlation with the labels is close to 0 for all components, supporting our assumption. 

\begin{figure}[ht]
\begin{center}
  \subfloat[\centering ImageNet]{{\label{imagenetspatial}\includegraphics[width=0.45\columnwidth]{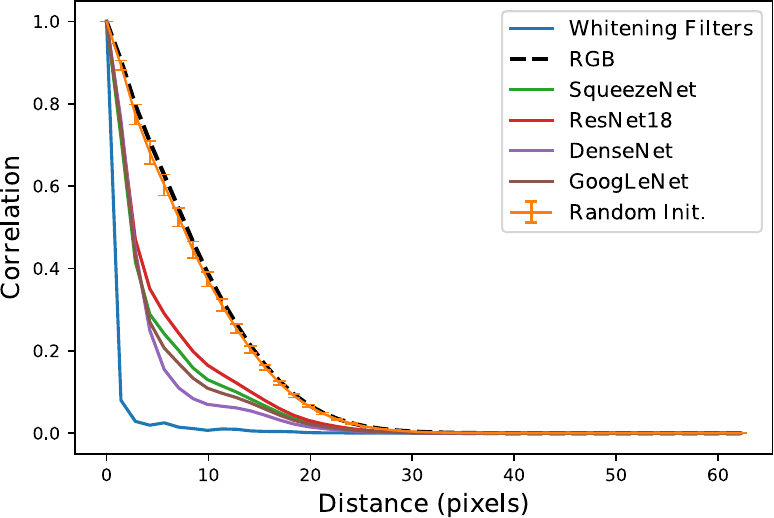} }}%
  \hfill
  \subfloat[\centering CIFAR10]{{\label{cifarspatial}\includegraphics[width=0.45\columnwidth]{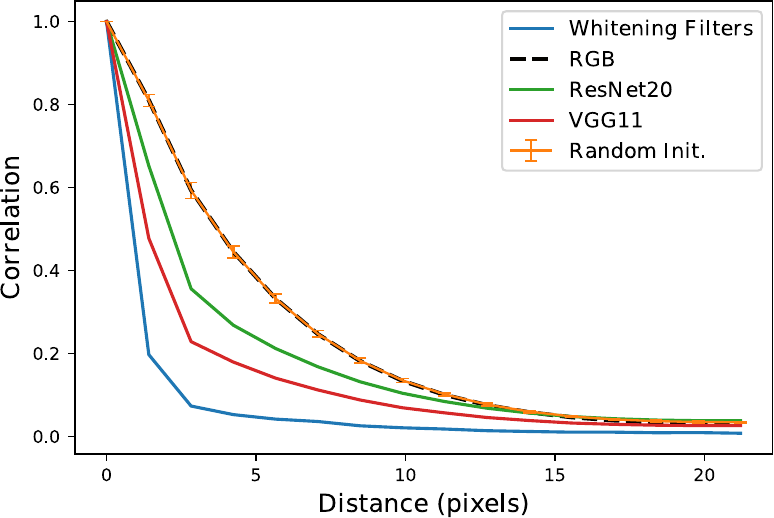}}}%
\caption{Auto-correlation as function of distance between activation maps of the first layer and RGB, for CIFAR10 and ImageNet. Learned filters in the first layer seem to spatially decorrelate pixels relative to RGB space, or a transformation with random filters. Note that standard deviation of the random initialization is hardly noticeable despite being averaged over 100 different random seeds, suggesting the discrepancy to trained models isn't coincidental.}
\label{autocorrelation}
\end{center}
\vskip 0.1in
\end{figure}

One prediction from our analysis is that the first layer of CNNs should perform partial decorrelation. Let $y$ be the vector that denotes the output of all channels at a particular location. The autocorrelation function is defined as $  C(\delta)= \E_i [ y(i) y(i+ \delta)]$
where the expectation is taken over locations $(i)$ and training images. We first compute this autocorrelation when $y$ includes three channels corresponding to the input (R,G,B). As can be seen in figure~\ref{autocorrelation} the correlation decreases as $\delta$ increases, but even at a distance of $10$ pixels the correlation is above $0.5$. When we measure this same autocorrelation function with random filters, the vector $y$ is of length 64, but the autocorrelation function is almost identical to that of RGB (note that the graph corresponding to random weights includes error bars and summarizes 100 different random initializations but all random initializations give very similar autocorrelation functions). In contrast, when the vector $y$ is the output of all 64 channels in a {\em learned} representation, we consistently find that the spatial correlation is significantly reduced, 
(e.g. at a distance of 10 pixels the correlation after learning is reduced to around $0.2$). For comparison, we also show the autocorrelation function of a set of filters that satisfy perfect whitening which reduces the correlation at distance 10 pixels to zero, as expected. Thus, consistent with our theoretical analysis of linear CNNs, real-world CNNs perform approximate whitening of the input and remove much of the redundancy that is present in their input even though they are not explicitly trained with a redundancy reduction loss. 

\begin{figure*}[t!]
  \centering
  \subfloat[\centering VGG11 on CIFAR10 (0.96)]{{\label{trueradomenergy1}\includegraphics[width=0.45\linewidth]{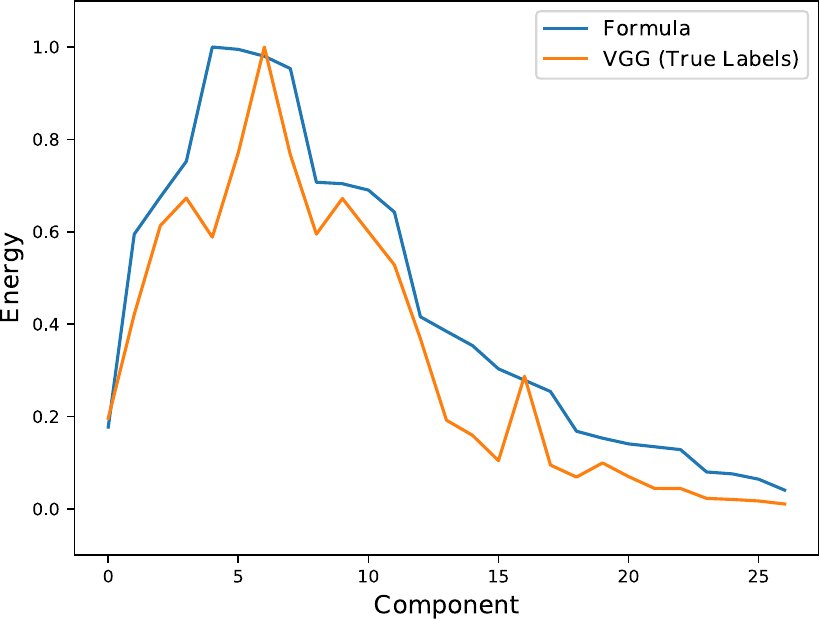} }}%
  \hfill
  \subfloat[\centering VGG11 on CelebA (0.97)]{{\label{trueradomenergy2}\includegraphics[width=0.45\linewidth]{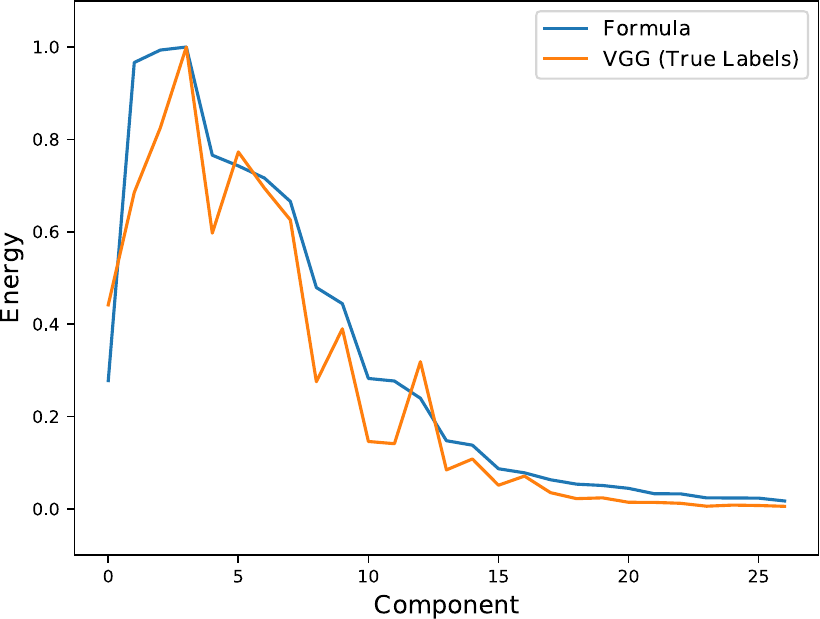} }}%
    \hfill
  \subfloat[\centering ResNet18 on Imagenet (0.92)]{{\label{trueradomenergy3}\includegraphics[width=0.45\linewidth]{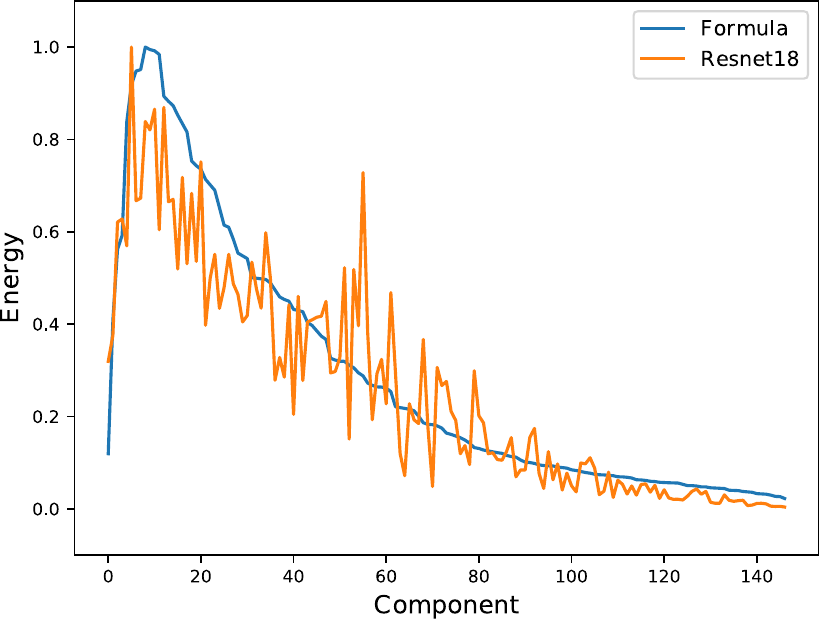} }}%
   
  \caption{Examples of fitting the formula to different networks trained on different datasets. Overall, the formula captures the trend learned in the first layer of the networks. Correlation coefficients in parentheses, and see \cref{table:formulafit} for more.}\label{fig:formulafits}
     \vskip 0.1in
\end{figure*}

Not only does our analysis predict this partial decorrelation of the input at a finite number of training iterations, it also gives a precise characterization of the energy profiles for a linear CNN. Does this formula predict the energy profiles of real models? We compare \cref{to-provew-eq-main} to energy profiles of real models by setting a constant learning rate for all datasets and searching over the number of gradient steps $t$. The results, portrayed in \cref{fig:formulafits} show \textbf{high correlation between the formula and real-world models} (consistently above 0.9), even in complex datasets such as ImageNet. \cref{table:formulafit} expands on these by providing correlation coefficients of the formula to different models, with multiple random seeds and on many datasets. Consistently, the formula calculated at a finite iteration is able to capture much of what is done by the first layer, independent of dataset, but not that of a random initalization. More fits are provided in \cref{appendix:formulafitexpanded}.

\begin{table}[h]
\caption{Correlation between energy profiles of VGG11 with the analytic formula for CIFAR10, CelebA and different binary datasets of CIFAR10, averaged over 3 different seeds. Correlations for ImageNet are averaged over 5 different models (and see \cref{appendix:formulafitexpanded}). The correlation with a random initialization is also presented for reference.}
\label{table:formulafit}
\vskip 0.15in
\begin{center}
\begin{small}
\begin{sc}

\begin{tabular}{lccccr}
\toprule
Dataset & Correlation \\

\midrule
ImageNet&0.9$\pm$0.01\\
Cifar10&0.94$\pm$0.01\\
CelebA&0.96$\pm$0.01\\
Car vs Truck&0.93$\pm$0.01\\
Dog vs Frog&0.91$\pm$0.02\\
Dog vs Cat&0.95$\pm$0.01\\
Bird vs Plane&0.96$\pm$0.005\\
Boat vs Plane&0.96$\pm$0.01\\
Random Init.&0.1$\pm$0.15\\

\bottomrule
\end{tabular}
\end{sc}
\end{small}
\end{center}
 \vskip 0.1in
\end{table}



\begin{figure}[h!]
  \centering
  \subfloat[\centering Labeling by the 15'th Component]{{\label{comp15}\includegraphics[width=0.45\columnwidth]{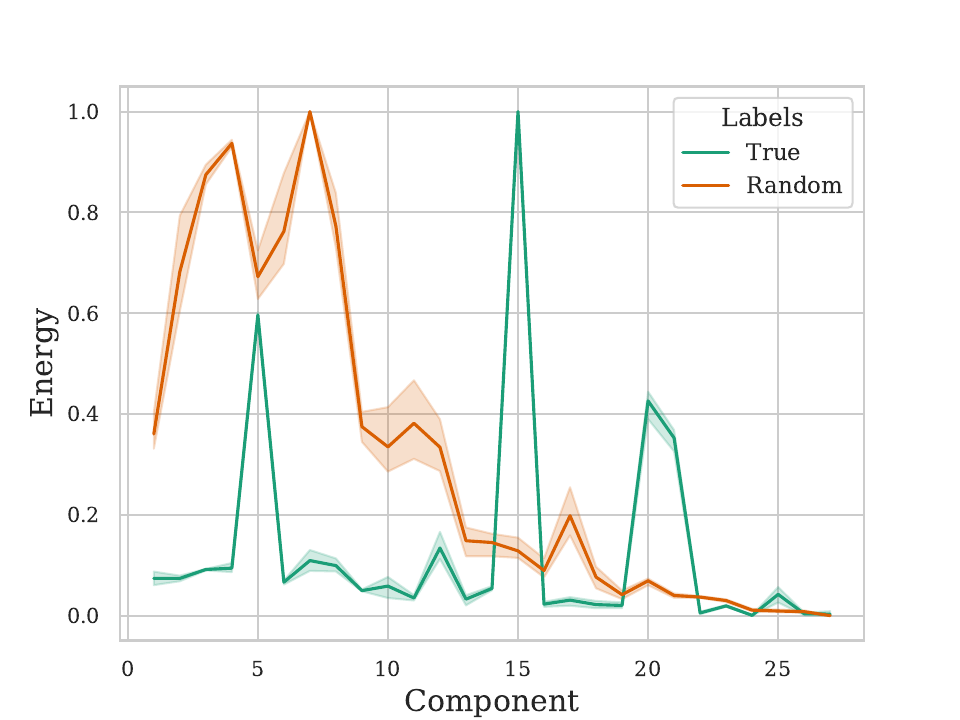} }}%
  \hfill
    \subfloat[\centering Enhancing the 25'th Component]{{\label{comp25enhance}\includegraphics[width=0.45\columnwidth]{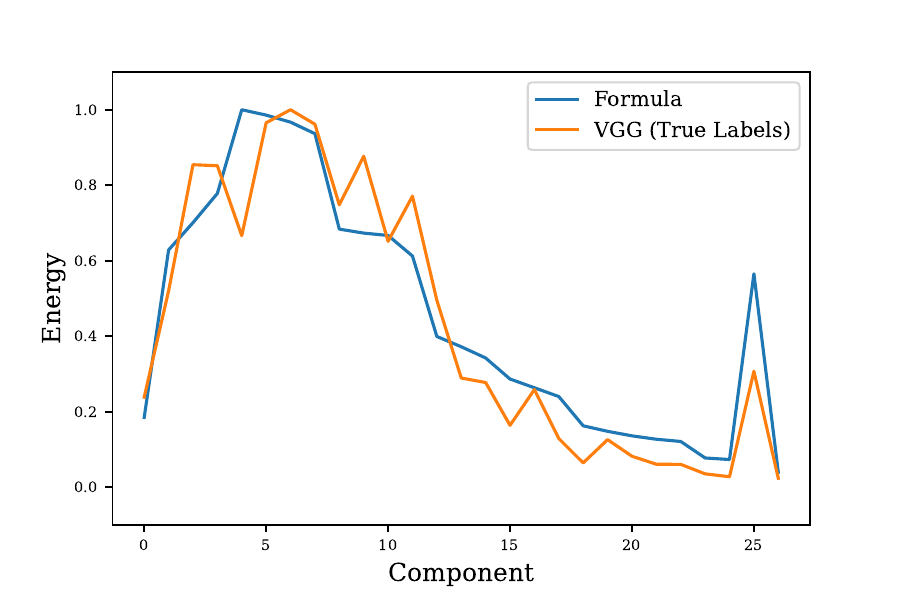} }}
  \caption{Changing the joint distribution of the patch energy and the labels affects the energy profiles. When correlation between the energy and the labels is introduced (\ref{comp15}) then the correlation between the true and random profiles is broken. When statistics are changed without introducing correlation (\ref{comp25enhance}) the theory follows suit. More results can be found in \cref{changingstats_appendix}}%
  \label{fig:componentlabeling}%
  \vskip 0.1in  
\end{figure}


In a final test of the ability of our analytical formula to fit the energy profiles of real, nonlinear CNNs we design two experiments that attempt to change the energy and label statistics of the classification task. In the first task we force the true labels to correlate with the input - for the $i$'th PCA component, we sort all CIFAR10 images by their average energy in the $i$'th direction and divide them into 10 equally sized sets. This creates 27 datasets (as the number of PCA components for $3\times 3 \times 3$ patches), each with high correlation between the labels and the image energy. \cref{comp15} displays the result of this experiment conducted on the 15'th component. As expected, the profiles of true and random labels are now noticeably different and their correlation drops to around 0 (and see \cref{appendix:labelpatchcor}), only by introducing correlation between the images' patch energy and labels.


In another experiment, we change only the input statistics by multiplying each patch by a constant factor $\alpha$ in a constant PCA direction. In this setting, there is no change in correlation between the patch energy and the labels as the same transformation is applied to all patches. Additionally, the eigenvalue corresponding the the component we enhanced is changed from $\lambda_i$ to $\alpha^2\lambda_i$. \cref{comp25enhance} shows that as expected, our analytic formula for random labels still captures the energy profiles of VGG with true labels, after applying the same transformation that was done to the input images to the eigenvalues used in the formula. More results are presented in \cref{appendix:eigentransform}.

\section{Related Works}
There have been many studies devoted to comparing representations in different neural networks  \citep{oldsimilarity, understandingequivalences,csis021similarity}.  The comparison is often done by comparing the output of transformations induced by the neurons \citep{CKA,minibatchCKA,doimo} or the neurons themselves \citep{subspacematching,bipartitematch}. The energy profile is an alternative method that is especially useful for comparing linear representations and avoids many of the pitfalls of previous approaches.  


The fact that different CNNs tend to learn qualitatively similar filters in the first layer has been reported previously \citep[for example]{bengio_transfer,gabor_cnn, gabor_cnn2, gabornet}, and follows from a line of work of visualizing representations in deep CNNs \cite{cnn_viz1,cnn_viz2}. Our work extends this finding by showing that the overall representation in the first layer is not only qualitatively but also is {\em quantitatively} similar - different CNNs not only learn to recognize spatial frequencies in their first layer but also the same distribution of frequencies. This consistency is then expanded to networks trained with true and random labels.


The idea that early representations should remove redundancies in their input goes back to \citet{barlow} and there has been a great deal of work arguing that initial layers in biological neural networks remove dependencies in their input~\cite{field1994goal,olshausen1996emergence,BELL19973327}. In particular, when explicit redundancy reduction is performed on natural image data, this principle leads to Gabor filters similar to those that are observed in the first layer of CNNs. In  this work we  followed~\citet{AtickRedlich90} and focused on removing linear dependencies by whitening. More importantly, we have shown that this form  of redundancy reduction emerges from minimizing the classification loss either with true or random labels. 

The usefulness of whitening as a normalization step in image processing techniques is well known \citep{natim}, and is even used as a preprocessing technique when training CNNs \cite{coateswhitening,cnnwhitening}. This has inspired others to constrain intermediate representations of neural networks to be white as well \citep{zca_batch_normalization,dropoutinsteadofwhitening,wnn,gwnn,Pan_2019_ICCV,Zhang_2021_CVPR} in order to improve convergence time and performance. Our work shows that approximate whitening occurs in CNNs even without an explicit whitening preprocessing step nor without an explicit "redundancy reduction" loss. 

As previously explained, the emergence of whitening is partially the result of a bias in the gradient descent training algorithm. The fact that gradient descent training biases towards certain solutions has been known for many years, and proven mainly for linear predictors and separable data. Studies on linear networks \citep{SoudryHNGS18} and linear CNNs \citep{gd_implicit_bias} found that under certain conditions, gradient descent causes the effective linear predictor to be biased towards sparsity (in Fourier space in the case of CNNs) or minimal norm or max-margin \citep{bach_implicit_bias}. Similar works have also shown that deep non-linear networks are biased towards learning lower frequencies first \citep{spectral_bias}. Our theoretical analysis follows this line, and that of gaining insight into real-world networks from simpler linear models \citep{lecun_dynamics,pcbias,gidel,gissin}, while verifying our claims by quantitatively showing consistency between theory and practice.  


Previous works have examined the usefulness of representations in models trained with random labels by incorporating them in transfer learning. Indeed, we show explicitly that since there is a high degree of similarity between the first layer of models trained with true labels and random ones, it is reasonable to assume that the layers of random models could be useful for transfer learning. While \citet{stitching} claimed that this was due to similarity between the first layer of a model trained with random labels and a random initialization, \citet{ganavim} offered the explanation that the first layer filters' covariance and the patch PCA have the same eigenvectors. Our results contradict the hypothesis of \citep{stitching} and extend the results of \citep{ganavim} to give an analytical formula for the energy profile that holds for true and random labels.

\section{Discussion}
The dramatic success of CNNs has lead to increased interest in the representations they learn, whether for explainability or usefulness in other tasks. In this paper we have focused on the representation that CNNs learn in the very first layer and presented a high degree of quantitative consistency between the energy profiles learned by different networks using different initializations, architectures, and even labels.  To understand why CNNs learn this particular energy profile we have analyzed linear CNNs and shown that this consistency is not a result of usefulness for object recognition but rather due to properties of the input and output statistics, namely no correlation between image patches and labels, and a bias of the training algorithm. Combined, the two give an implicit bias towards partial "redundancy reduction".

To generalize to real-world CNNs, we showed that the analytic formulation of the linear case captures much of what is done by the first layer of different networks on different datasets. To complement our explanation, we designed experiments that adjust the statistics of the input and output and showed the results behave as predicted. 

Redundancy reduction is closely related to what is commonly referred to as "disentanglement" in deep learning~\cite{goodfellow2016deep}:  representations of the input should disentangle the different factors of variation that influence each piece of the input. Our results shows that real-world CNNs trained with gradient descent perform a simplified version of disentanglement even if there is no explicit loss that rewards it. It will be interesting to see if this result can be extended to deeper layers and more nonlinear definitions of disentanglement. 
\section{Acknowledgements}
We would like the thank the Israeli Ministry of Science and Technology and the Gatsby Foundation for their support in funding this research.

\FloatBarrier
\newpage
\bibliography{icml2023}

\begin{thebibliography}{44}
\providecommand{\natexlab}[1]{#1}
\providecommand{\url}[1]{\texttt{#1}}
\expandafter\ifx\csname urlstyle\endcsname\relax
  \providecommand{\doi}[1]{doi: #1}\else
  \providecommand{\doi}{doi: \begingroup \urlstyle{rm}\Url}\fi

\bibitem[Alekseev \& Bobe(2019)Alekseev and Bobe]{gabornet}
Alekseev, A. and Bobe, A.
\newblock Gabornet: Gabor filters with learnable parameters in deep
  convolutional neural network.
\newblock In \emph{2019 International Conference on Engineering and
  Telecommunication (EnT)}, pp.\  1--4, 2019.
\newblock \doi{10.1109/EnT47717.2019.9030571}.

\bibitem[Arpit et~al.(2017)Arpit, Jastrzebski, Ballas, Krueger, Bengio, Kanwal,
  Maharaj, Fischer, Courville, Bengio, and Lacoste-Julien]{memorization}
Arpit, D., Jastrzebski, S., Ballas, N., Krueger, D., Bengio, E., Kanwal, M.~S.,
  Maharaj, T., Fischer, A., Courville, A., Bengio, Y., and Lacoste-Julien, S.
\newblock A closer look at memorization in deep networks.
\newblock In Precup, D. and Teh, Y.~W. (eds.), \emph{Proceedings of the 34th
  International Conference on Machine Learning}, volume~70 of \emph{Proceedings
  of Machine Learning Research}, pp.\  233--242. PMLR, 06--11 Aug 2017.
\newblock URL \url{https://proceedings.mlr.press/v70/arpit17a.html}.

\bibitem[Atick \& Redlich(1990)Atick and Redlich]{AtickRedlich90}
Atick, J.~J. and Redlich, A.~N.
\newblock {Towards a Theory of Early Visual Processing}.
\newblock \emph{Neural Computation}, 2\penalty0 (3):\penalty0 308--320, 09
  1990.
\newblock ISSN 0899-7667.
\newblock \doi{10.1162/neco.1990.2.3.308}.
\newblock URL \url{https://doi.org/10.1162/neco.1990.2.3.308}.

\bibitem[Bansal et~al.(2021)Bansal, Nakkiran, and Barak]{stitching}
Bansal, Y., Nakkiran, P., and Barak, B.
\newblock Revisiting model stitching to compare neural representations.
\newblock In Ranzato, M., Beygelzimer, A., Dauphin, Y., Liang, P., and Vaughan,
  J.~W. (eds.), \emph{Advances in Neural Information Processing Systems},
  volume~34, pp.\  225--236. Curran Associates, Inc., 2021.
\newblock URL
  \url{https://proceedings.neurips.cc/paper/2021/file/01ded4259d101feb739b06c399e9cd9c-Paper.pdf}.

\bibitem[Barlow(1989)]{barlow}
Barlow, H.
\newblock {Unsupervised Learning}.
\newblock \emph{Neural Computation}, 1\penalty0 (3):\penalty0 295--311, 09
  1989.
\newblock ISSN 0899-7667.
\newblock \doi{10.1162/neco.1989.1.3.295}.
\newblock URL \url{https://doi.org/10.1162/neco.1989.1.3.295}.

\bibitem[Bell \& Sejnowski(1997)Bell and Sejnowski]{BELL19973327}
Bell, A.~J. and Sejnowski, T.~J.
\newblock The “independent components” of natural scenes are edge filters.
\newblock \emph{Vision Research}, 37\penalty0 (23):\penalty0 3327--3338, 1997.
\newblock ISSN 0042-6989.
\newblock \doi{https://doi.org/10.1016/S0042-6989(97)00121-1}.
\newblock URL
  \url{https://www.sciencedirect.com/science/article/pii/S0042698997001211}.

\bibitem[Chen et~al.(2019)Chen, Chen, Shi, Hsieh, Liao, and
  Zhang]{dropoutinsteadofwhitening}
Chen, G., Chen, P., Shi, Y., Hsieh, C., Liao, B., and Zhang, S.
\newblock Rethinking the usage of batch normalization and dropout in the
  training of deep neural networks.
\newblock \emph{CoRR}, abs/1905.05928, 2019.
\newblock URL \url{http://arxiv.org/abs/1905.05928}.

\bibitem[Chizat \& Bach(2020)Chizat and Bach]{bach_implicit_bias}
Chizat, L. and Bach, F.
\newblock Implicit bias of gradient descent for wide two-layer neural networks
  trained with the logistic loss.
\newblock In Abernethy, J. and Agarwal, S. (eds.), \emph{Proceedings of Thirty
  Third Conference on Learning Theory}, volume 125 of \emph{Proceedings of
  Machine Learning Research}, pp.\  1305--1338. PMLR, 09--12 Jul 2020.
\newblock URL \url{https://proceedings.mlr.press/v125/chizat20a.html}.

\bibitem[Coates et~al.(2011)Coates, Ng, and Lee]{coateswhitening}
Coates, A., Ng, A., and Lee, H.
\newblock An analysis of single-layer networks in unsupervised feature
  learning.
\newblock In Gordon, G., Dunson, D., and Dudík, M. (eds.), \emph{Proceedings
  of the Fourteenth International Conference on Artificial Intelligence and
  Statistics}, volume~15 of \emph{Proceedings of Machine Learning Research},
  pp.\  215--223, Fort Lauderdale, FL, USA, 11--13 Apr 2011. PMLR.
\newblock URL \url{https://proceedings.mlr.press/v15/coates11a.html}.

\bibitem[Csiszárik et~al.(2021)Csiszárik, Kőrösi-Szabó, Ákos
  K.~Matszangosz, Papp, and Varga]{csis021similarity}
Csiszárik, A., Kőrösi-Szabó, P., Ákos K.~Matszangosz, Papp, G., and Varga,
  D.
\newblock Similarity and matching of neural network representations, 2021.

\bibitem[Desjardins et~al.(2015)Desjardins, Simonyan, Pascanu, and
  kavukcuoglu]{wnn}
Desjardins, G., Simonyan, K., Pascanu, R., and kavukcuoglu, k.
\newblock Natural neural networks.
\newblock In Cortes, C., Lawrence, N., Lee, D., Sugiyama, M., and Garnett, R.
  (eds.), \emph{Advances in Neural Information Processing Systems}, volume~28.
  Curran Associates, Inc., 2015.
\newblock URL
  \url{https://proceedings.neurips.cc/paper/2015/file/2de5d16682c3c35007e4e92982f1a2ba-Paper.pdf}.

\bibitem[Ding et~al.(2021)Ding, Denain, and Steinhardt]{ding2021grounding}
Ding, F., Denain, J.-S., and Steinhardt, J.
\newblock Grounding representation similarity with statistical testing.
\newblock \emph{arXiv preprint arXiv:2108.01661}, 2021.

\bibitem[Doimo et~al.(2020)Doimo, Glielmo, Ansuini, and Laio]{doimo}
Doimo, D., Glielmo, A., Ansuini, A., and Laio, A.
\newblock Hierarchical nucleation in deep neural networks.
\newblock In \emph{Proceedings of the 34th International Conference on Neural
  Information Processing Systems}, NIPS'20, Red Hook, NY, USA, 2020. Curran
  Associates Inc.
\newblock ISBN 9781713829546.

\bibitem[Field(1994)]{field1994goal}
Field, D.~J.
\newblock What is the goal of sensory coding?
\newblock \emph{Neural computation}, 6\penalty0 (4):\penalty0 559--601, 1994.

\bibitem[Gidaris et~al.(2018)Gidaris, Singh, and
  Komodakis]{predicting_rotations}
Gidaris, S., Singh, P., and Komodakis, N.
\newblock Unsupervised representation learning by predicting image rotations.
\newblock \emph{CoRR}, abs/1803.07728, 2018.
\newblock URL \url{http://arxiv.org/abs/1803.07728}.

\bibitem[Gidel et~al.(2019)Gidel, Bach, and Lacoste-Julien]{gidel}
Gidel, G., Bach, F., and Lacoste-Julien, S.
\newblock Implicit regularization of discrete gradient dynamics in linear
  neural networks.
\newblock In Wallach, H., Larochelle, H., Beygelzimer, A., d\textquotesingle
  Alch\'{e}-Buc, F., Fox, E., and Garnett, R. (eds.), \emph{Advances in Neural
  Information Processing Systems}, volume~32. Curran Associates, Inc., 2019.
\newblock URL
  \url{https://proceedings.neurips.cc/paper/2019/file/f39ae9ff3a81f499230c4126e01f421b-Paper.pdf}.

\bibitem[Girshick et~al.(2013)Girshick, Donahue, Darrell, and Malik]{cnn_viz2}
Girshick, R.~B., Donahue, J., Darrell, T., and Malik, J.
\newblock Rich feature hierarchies for accurate object detection and semantic
  segmentation.
\newblock \emph{CoRR}, abs/1311.2524, 2013.
\newblock URL \url{http://arxiv.org/abs/1311.2524}.

\bibitem[Gissin et~al.(2019)Gissin, Shalev{-}Shwartz, and Daniely]{gissin}
Gissin, D., Shalev{-}Shwartz, S., and Daniely, A.
\newblock The implicit bias of depth: How incremental learning drives
  generalization.
\newblock \emph{CoRR}, abs/1909.12051, 2019.
\newblock URL \url{http://arxiv.org/abs/1909.12051}.

\bibitem[Goodfellow et~al.(2016)Goodfellow, Bengio, and
  Courville]{goodfellow2016deep}
Goodfellow, I., Bengio, Y., and Courville, A.
\newblock \emph{Deep learning}.
\newblock MIT press, 2016.

\bibitem[Gunasekar et~al.(2018)Gunasekar, Lee, Soudry, and
  Srebro]{gd_implicit_bias}
Gunasekar, S., Lee, J.~D., Soudry, D., and Srebro, N.
\newblock Implicit bias of gradient descent on linear convolutional networks.
\newblock In Bengio, S., Wallach, H., Larochelle, H., Grauman, K.,
  Cesa-Bianchi, N., and Garnett, R. (eds.), \emph{Advances in Neural
  Information Processing Systems}, volume~31. Curran Associates, Inc., 2018.
\newblock URL
  \url{https://proceedings.neurips.cc/paper/2018/file/0e98aeeb54acf612b9eb4e48a269814c-Paper.pdf}.

\bibitem[Hacohen \& Weinshall(2022)Hacohen and Weinshall]{pcbias}
Hacohen, G. and Weinshall, D.
\newblock Principal components bias in over-parameterized linear models, and
  its manifestation in deep neural networks.
\newblock \emph{Journal of Machine Learning Research}, 23\penalty0
  (155):\penalty0 1--46, 2022.
\newblock URL \url{http://jmlr.org/papers/v23/21-0991.html}.

\bibitem[Huang et~al.(2018)Huang, Yang, Lang, and
  Deng]{zca_batch_normalization}
Huang, L., Yang, D., Lang, B., and Deng, J.
\newblock Decorrelated batch normalization.
\newblock In \emph{Proceedings of the IEEE Conference on Computer Vision and
  Pattern Recognition (CVPR)}, June 2018.

\bibitem[Hyvärinen et~al.(2009)Hyvärinen, Hurri, and Hoyer]{natim}
Hyvärinen, A., Hurri, J., and Hoyer, P.~O. (eds.).
\newblock \emph{Natural Image Statistics}.
\newblock Springer-Verlag London, London, UK, 2009.

\bibitem[Kornblith et~al.(2019)Kornblith, Norouzi, Lee, and Hinton]{CKA}
Kornblith, S., Norouzi, M., Lee, H., and Hinton, G.
\newblock Similarity of neural network representations revisited.
\newblock In Chaudhuri, K. and Salakhutdinov, R. (eds.), \emph{Proceedings of
  the 36th International Conference on Machine Learning}, volume~97 of
  \emph{Proceedings of Machine Learning Research}, pp.\  3519--3529. PMLR,
  09--15 Jun 2019.
\newblock URL \url{https://proceedings.mlr.press/v97/kornblith19a.html}.

\bibitem[Krizhevsky et~al.(2012)Krizhevsky, Sutskever, and Hinton]{AlexNet}
Krizhevsky, A., Sutskever, I., and Hinton, G.~E.
\newblock Imagenet classification with deep convolutional neural networks.
\newblock In Pereira, F., Burges, C., Bottou, L., and Weinberger, K. (eds.),
  \emph{Advances in Neural Information Processing Systems}, volume~25. Curran
  Associates, Inc., 2012.
\newblock URL
  \url{https://proceedings.neurips.cc/paper/2012/file/c399862d3b9d6b76c8436e924a68c45b-Paper.pdf}.

\bibitem[Laakso \& Cottrell(2000)Laakso and Cottrell]{oldsimilarity}
Laakso, A. and Cottrell, G.
\newblock Content and cluster analysis: Assessing representational similarity
  in neural systems.
\newblock \emph{Philosophical Psychology}, 13\penalty0 (1):\penalty0 47--76,
  2000.
\newblock \doi{10.1080/09515080050002726}.
\newblock URL \url{https://doi.org/10.1080/09515080050002726}.

\bibitem[LeCun et~al.(1991)LeCun, Kanter, and Solla]{lecun_dynamics}
LeCun, Y., Kanter, I., and Solla, S.
\newblock Second order properties of error surfaces: Learning time and
  generalization.
\newblock In Lippmann, R.~P., Moody, J., and Touretzky, D. (eds.),
  \emph{Advances in Neural Information Processing Systems}, volume~3.
  Morgan-Kaufmann, 1991.
\newblock URL
  \url{https://proceedings.neurips.cc/paper/1990/file/758874998f5bd0c393da094e1967a72b-Paper.pdf}.

\bibitem[Lenc \& Vedaldi(2015)Lenc and Vedaldi]{understandingequivalences}
Lenc, K. and Vedaldi, A.
\newblock Understanding image representations by measuring their equivariance
  and equivalence.
\newblock In \emph{2015 IEEE Conference on Computer Vision and Pattern
  Recognition (CVPR)}, pp.\  991--999, Los Alamitos, CA, USA, jun 2015. IEEE
  Computer Society.
\newblock \doi{10.1109/CVPR.2015.7298701}.
\newblock URL
  \url{https://doi.ieeecomputersociety.org/10.1109/CVPR.2015.7298701}.

\bibitem[Li et~al.(2015)Li, Yosinski, Clune, Lipson, and
  Hopcroft]{bipartitematch}
Li, Y., Yosinski, J., Clune, J., Lipson, H., and Hopcroft, J.
\newblock Convergent learning: Do different neural networks learn the same
  representations?
\newblock In Storcheus, D., Rostamizadeh, A., and Kumar, S. (eds.),
  \emph{Proceedings of the 1st International Workshop on Feature Extraction:
  Modern Questions and Challenges at NIPS 2015}, volume~44 of \emph{Proceedings
  of Machine Learning Research}, pp.\  196--212, Montreal, Canada, 11 Dec 2015.
  PMLR.
\newblock URL \url{https://proceedings.mlr.press/v44/li15convergent.html}.

\bibitem[Luan et~al.(2017)Luan, Zhang, Chen, Cao, Han, and Liu]{gabor_cnn2}
Luan, S., Zhang, B., Chen, C., Cao, X., Han, J., and Liu, J.
\newblock Gabor convolutional networks.
\newblock \emph{CoRR}, abs/1705.01450, 2017.
\newblock URL \url{http://arxiv.org/abs/1705.01450}.

\bibitem[Luo(2017)]{gwnn}
Luo, P.
\newblock Learning deep architectures via generalized whitened neural networks.
\newblock In Precup, D. and Teh, Y.~W. (eds.), \emph{Proceedings of the 34th
  International Conference on Machine Learning}, volume~70 of \emph{Proceedings
  of Machine Learning Research}, pp.\  2238--2246. PMLR, 06--11 Aug 2017.
\newblock URL \url{https://proceedings.mlr.press/v70/luo17a.html}.

\bibitem[Maennel et~al.(2020)Maennel, Alabdulmohsin, Tolstikhin, Baldock,
  Bousquet, Gelly, and Keysers]{ganavim}
Maennel, H., Alabdulmohsin, I., Tolstikhin, I., Baldock, R. J.~N., Bousquet,
  O., Gelly, S., and Keysers, D.
\newblock What do neural networks learn when trained with random labels?
\newblock 2020.
\newblock URL \url{https://arxiv.org/abs/2006.10455}.

\bibitem[Nguyen et~al.(2021)Nguyen, Raghu, and Kornblith]{minibatchCKA}
Nguyen, T., Raghu, M., and Kornblith, S.
\newblock Do wide and deep networks learn the same things? uncovering how
  neural network representations vary with width and depth.
\newblock In \emph{International Conference on Learning Representations}, 2021.
\newblock URL \url{https://openreview.net/forum?id=KJNcAkY8tY4}.

\bibitem[Olshausen \& Field(1996)Olshausen and Field]{olshausen1996emergence}
Olshausen, B.~A. and Field, D.~J.
\newblock Emergence of simple-cell receptive field properties by learning a
  sparse code for natural images.
\newblock \emph{Nature}, 381\penalty0 (6583):\penalty0 607--609, 1996.

\bibitem[Owsley(2003)]{CSF}
Owsley, C.
\newblock Contrast sensitivity.
\newblock \emph{Ophthalmology clinics of North America}, 16\penalty0
  (2):\penalty0 171—177, June 2003.
\newblock ISSN 0896-1549.
\newblock \doi{10.1016/s0896-1549(03)00003-8}.
\newblock URL \url{https://doi.org/10.1016/s0896-1549(03)00003-8}.

\bibitem[Pal \& Sudeep(2016)Pal and Sudeep]{cnnwhitening}
Pal, K.~K. and Sudeep, K.~S.
\newblock Preprocessing for image classification by convolutional neural
  networks.
\newblock In \emph{2016 IEEE International Conference on Recent Trends in
  Electronics, Information \& Communication Technology (RTEICT)}, pp.\
  1778--1781, 2016.
\newblock \doi{10.1109/RTEICT.2016.7808140}.

\bibitem[Pan et~al.(2019)Pan, Zhan, Shi, Tang, and Luo]{Pan_2019_ICCV}
Pan, X., Zhan, X., Shi, J., Tang, X., and Luo, P.
\newblock Switchable whitening for deep representation learning.
\newblock In \emph{Proceedings of the IEEE/CVF International Conference on
  Computer Vision (ICCV)}, October 2019.

\bibitem[Rahaman et~al.(2019)Rahaman, Baratin, Arpit, Draxler, Lin, Hamprecht,
  Bengio, and Courville]{spectral_bias}
Rahaman, N., Baratin, A., Arpit, D., Draxler, F., Lin, M., Hamprecht, F.,
  Bengio, Y., and Courville, A.
\newblock On the spectral bias of neural networks.
\newblock In Chaudhuri, K. and Salakhutdinov, R. (eds.), \emph{Proceedings of
  the 36th International Conference on Machine Learning}, volume~97 of
  \emph{Proceedings of Machine Learning Research}, pp.\  5301--5310. PMLR,
  09--15 Jun 2019.
\newblock URL \url{https://proceedings.mlr.press/v97/rahaman19a.html}.

\bibitem[Sarwar et~al.(2017)Sarwar, Panda, and Roy]{gabor_cnn}
Sarwar, S.~S., Panda, P., and Roy, K.
\newblock Gabor filter assisted energy efficient fast learning convolutional
  neural networks.
\newblock \emph{CoRR}, abs/1705.04748, 2017.
\newblock URL \url{http://arxiv.org/abs/1705.04748}.

\bibitem[Soudry et~al.(2018)Soudry, Hoffer, and Srebro]{SoudryHNGS18}
Soudry, D., Hoffer, E., and Srebro, N.
\newblock The implicit bias of gradient descent on separable data.
\newblock In \emph{International Conference on Learning Representations}, 2018.
\newblock URL \url{https://openreview.net/forum?id=r1q7n9gAb}.

\bibitem[Wang et~al.(2018)Wang, Hu, Gu, Hu, Wu, He, and
  Hopcroft]{subspacematching}
Wang, L., Hu, L., Gu, J., Hu, Z., Wu, Y., He, K., and Hopcroft, J.
\newblock Towards understanding learning representations: To what extent do
  different neural networks learn the same representation.
\newblock In Bengio, S., Wallach, H., Larochelle, H., Grauman, K.,
  Cesa-Bianchi, N., and Garnett, R. (eds.), \emph{Advances in Neural
  Information Processing Systems}, volume~31. Curran Associates, Inc., 2018.
\newblock URL
  \url{https://proceedings.neurips.cc/paper/2018/file/5fc34ed307aac159a30d81181c99847e-Paper.pdf}.

\bibitem[Yosinski et~al.(2014)Yosinski, Clune, Bengio, and
  Lipson]{bengio_transfer}
Yosinski, J., Clune, J., Bengio, Y., and Lipson, H.
\newblock How transferable are features in deep neural networks?
\newblock In Ghahramani, Z., Welling, M., Cortes, C., Lawrence, N., and
  Weinberger, K. (eds.), \emph{Advances in Neural Information Processing
  Systems}, volume~27. Curran Associates, Inc., 2014.
\newblock URL
  \url{https://proceedings.neurips.cc/paper/2014/file/375c71349b295fbe2dcdca9206f20a06-Paper.pdf}.

\bibitem[Zeiler \& Fergus(2013)Zeiler and Fergus]{cnn_viz1}
Zeiler, M.~D. and Fergus, R.
\newblock Visualizing and understanding convolutional networks.
\newblock \emph{CoRR}, abs/1311.2901, 2013.
\newblock URL \url{http://arxiv.org/abs/1311.2901}.

\bibitem[Zhang et~al.(2021)Zhang, Nezhadarya, Fashandi, Liu, Graham, and
  Shah]{Zhang_2021_CVPR}
Zhang, S., Nezhadarya, E., Fashandi, H., Liu, J., Graham, D., and Shah, M.
\newblock Stochastic whitening batch normalization.
\newblock In \emph{Proceedings of the IEEE/CVF Conference on Computer Vision
  and Pattern Recognition (CVPR)}, pp.\  10978--10987, June 2021.

\end{thebibliography}
\bibliographystyle{icml2023}

\newpage
\appendix
\onecolumn
\section{Proofs of Theorems on CNNs}\label{appendix:proof}

We start with some definitions. To simplify the notation, we assume
that the mean of the patches in the training set is zero. 

\begin{definition}
Given a set of patches $\{p_n\}$ the PCA vectors $u_i$ are eigenvectors of the matrix $\sum_{n} p_n p_n^T$. 
\end{definition}
\begin{definition}Given a set of filters $\{w_k\}$ and a set of PCA
vectors $\{u_i\}$ the energy profile
of the set is given by a vector $e$ whose $i$th component is given by:
\BE
e_i^2 = \frac{1}{K} \sum_{k=1}^K (w_k^T u_i)^2
\EE
\end{definition}
\begin{definition}Given a set of patches $\{p_n\}$ and a set of PCA
vectors $\{u_i\}$ the energy profile
of the set is given by a vector $\lambda$ whose $i$th component is given by:
\BE
\lambda_i^2 = \frac{1}{N} \sum_{n=1}^N (p_n^T u_i)^2
\EE
\end{definition}
\begin{definition}A labeled training set of images $\{x_n,y_n\}$
satisfies the property that the label is uncorrelated with individual
PCA coefficients if $E \left[ u_i^T p_j(x) y(x) \right] = E \left[
  u_i^T p_j(x) \right] E \left[y(x)) \right]$ where the expectation is over the dataset and $p_j(x)$ is a
randomly chosen patch in image $x$.
\end{definition}
\begin{theorem}
Consider a depth-2 linear CNN of any width
initialized with zero mean filters and variance $\sigma^2 I$ and trained
with gradient descent with step size $\eta$ on the MSE loss. Assume that different patches
in each image are uncorrelated with each other and that the labels are
uncorrelated with individual PCA components, then as the number of
patches in the training set goes to infinity, the energy profile of
the filters at iterations $t$ is given by: 

\begin{equation}
  \label{to-provew-eq}
  e_i  = \tilde{c}\cdot \frac{|1-(1-\eta \lambda_i^2)^t|}
    {\eta^2 \lambda_i^2} \lambda_i + \xi_{i}
    \end{equation}
where $\lambda_i$ is the energy profile of the training patches and
$\xi$ a random vector that depends on the initialization and whose
magnitude goes to zero as $\sigma \rightarrow 0$ .
\end{theorem}
\begin{proof}
The output of the network for an input image $x$ is given by:
\BE
\hat{y}(x) =  \sum_k \frac{1}{J}  \sum_{j=1}^J p_j(x)^T w_k = c  \bar{p}^T(x) \bar{w} 
\EE

where $p_j(x)$ is the $j$th patch in image $x$,  $\bar{p}(x)$ is the average patch in image $x$ and $\bar{w}$ is
the average filter and $c$ is the number of filters. This also means that the gradient of the MSE loss $L=\frac{1}{N} \sum_x (y(x)- \hat{y}(x))^2$  
with respect to a particular filter is given by:

\BE
\frac{\partial L}{\partial w_k} = c  \left( A \bar{w} - b \right)
\EE
where $A = \frac{1}{N} \sum_x \bar{p}(x) \bar{p}(x)^T$ and $b =
\frac{1}{N} \sum_x \bar{p}(x)y(x)$. Note that the gradient is the same
for all $k$ which means that at each iteration:
\BE
w_k(t)= \bar{w}(t) + w_k(0)
\EE

and we can describe the dynamics of the mean
filter at each iteration $t$ by: 

\BE
\bar{w}(t) = \bar{w}(t-1) - \eta  \left( A \bar{w}(t-1) - b \right)
\EE

Defining the matrix $C=(I-\eta A)$ and assuming that the mean filter
at the initial iteration is $0$ gives: 
\BE
\bar{w}(t)= \left( \sum_{n=0}^{t-1} C^n \right) b
\EE

Note that the matrix $C$ is diagonalized by the PCA basis and its
eigenvalues are $1-\eta \lambda_i$ which means that: 

\BE
u_i^T \bar{w}(t) = \frac{\left(1-(1-\eta \lambda_i^2)^t\right)}
{\eta^2 \lambda_i^2} (u_i^T b)
\EE

Or taking the absolute value of both sides:
\BE
\label{almost-eq}
|u_i^T \bar{w}(t)| = |\frac{\left(1-(1-\eta \lambda_i^2)^t\right)}
{\eta^2 \lambda_i^2}|\cdot |(u_i^T b)|
\EE

Now consider the term $|u_i^T b|$ this can be rewritten:
\BE
|u_i^T b| = \abs{ \frac{1}{N} \sum_x u_i^T \bar{p}(x) y(x)}
\EE

By the central limit theorem, the term $z_i=\frac{1}{N} \sum_x u_i^T
\bar{p}(x) y(x)$ approaches a Gaussian whose mean is the mean of the
random variable $ y (u_i^T p)$, i.e. the random variable is the
product of the label of an image and a PCA coefficient of the average patch in
that image. Since we are assuming the labels to be uncorrelated with
the PCA coefficient, the mean of this random variable is $0$ and its
variance is $\lambda_i^2/J$ (where $J$ is the number of patches).
Thus $z_i$ is a Gaussian random variable 
with mean zero and variance $\lambda_i^2/(JN)$ and the term $|u_i^T b|$ is a ``folded Gaussian'' whose expectation is: 
\BE
E(|u_i^T b|) =  \frac{{\lambda_i}}{\sqrt{JN}}
\frac{\sqrt{2}}{\sqrt{\pi}}
\EE
and whose variance is also proportional to $1/JN$. As $JN \rightarrow \infty$,
the variance goes to zero which means that $|u_i^T b|$ is
with high probability close to its expected value and hence $|u_i^T b|$
is with high probability proportional to $\lambda_i$. 

Substituting this in equation~\ref{almost-eq} gives that with high
probability:
\BE
\label{almost-eq2}
|u_i^T \bar{w}(t)| = c_2 \abs{\frac{\left(1-(1-\eta \lambda_i^2)^t\right)}
{\eta^2 \lambda_i^2}} \lambda_i
\EE

Finally, by the definition of the energy profile and the fact that
$w_k(t) = \bar{w}(t) + w_k(0)$ equation~\ref{to-provew-eq} follows.
\end{proof}

\begin{theorem}Let $\{w_k\}$ be the filters in the first layer of a
CNN. If the energy profile of these filters satisfy
equation~\ref{to-provew-eq} then as the number of iterations goes to infinity, the
filters in the first layer of the CNN perform spatial decorrelation. 
\end{theorem}
\begin{proof}
It is evident from equation~\ref{to-provew-eq} that as $t
\rightarrow \infty$, the energy profile is proportional to
$\frac{1}{\lambda}$. This means that the filter bank performs
``whitening'' and there have been many works that show the connection
of whitening to spatial decorrelation (see~\citet{natim} and
references within). For completeness, we give the derivation here. 

Recall that the PCA vectors of natural image patches aare
approximately the Fourier basis. Thus the fact that the energy profile
is proportional to $\frac{1}{\lambda}$ implies a relationship between
the Fourier transform of the bank of filters and the Fourier transform
of the images. Denote by $E(|X^F{\omega}|)$ the expected power spectrum
of the training images and by $|Wk^F(\omega)$ the power spectrum of
the $k$th filter then:
\BE
\label{whitening-eq}
\sum_k |w_k^{F}(\omega)|^2 \propto \frac{1}{E( |x^{F}(\omega)|^2)}
\EE

Now denote by $C$ the auto-correlation function of the representation
and by $y_k$ the $k$th channel, i.e. $y_k = x \star w_k$ then:
\BE
C = E( \sum_k  y_k \star y_k)
\EE
where the expectation is over images in the training set. We say that
a representation is ``spatially disentangled'' if the channels at
different locations are uncorrelated and $C$ is a delta function.

We denote by $C^F(\omega)$
the Fourier Transform of $C$ and $y_k^{F}(\omega)$ are the Fourier
transforms of each channel. Then:
\BEA
C^{F}(\omega) &=& E (\sum_k |y_k^{F}(\omega)|^2) \\
&=& \sum_k |w_k^{F}(\omega)|^2 E(|x^{F}(\omega)|^2) \\
&=&  E |x^{F}(\omega)|^2 \sum_k |w_k^{F}(\omega)|^2  \\
&=& c
\EEA
Where the last equation is derived by substituting
equation~\ref{whitening-eq}. Hence the Fourier Transform of the
auto-correlation function is a constant which means that the
auto-correlation function is a $\delta$ function.
\end{proof}

\section{Consistency for Different Datasets and Architectures}\label{appendix:pretrainedcors}
\begin{figure*}[h]
  \centering
  \subfloat[\centering  Energy Profiles]{{\includegraphics[width=0.52\linewidth]{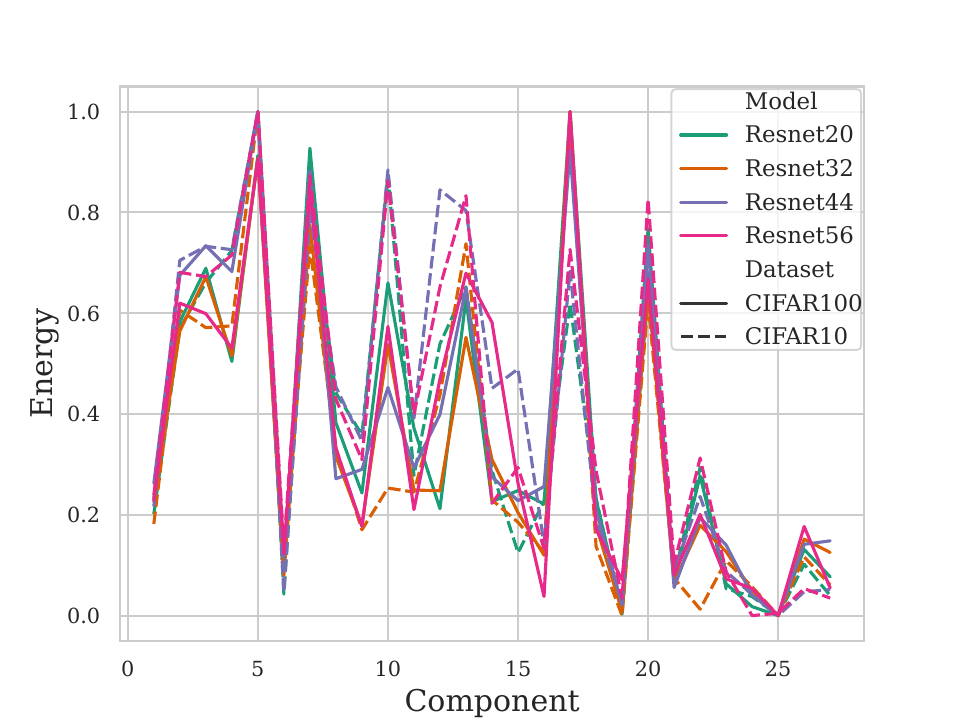} }}%
  \hfill
  \subfloat[\centering  Correlation Coefficients]{{\includegraphics[width=0.40\linewidth]{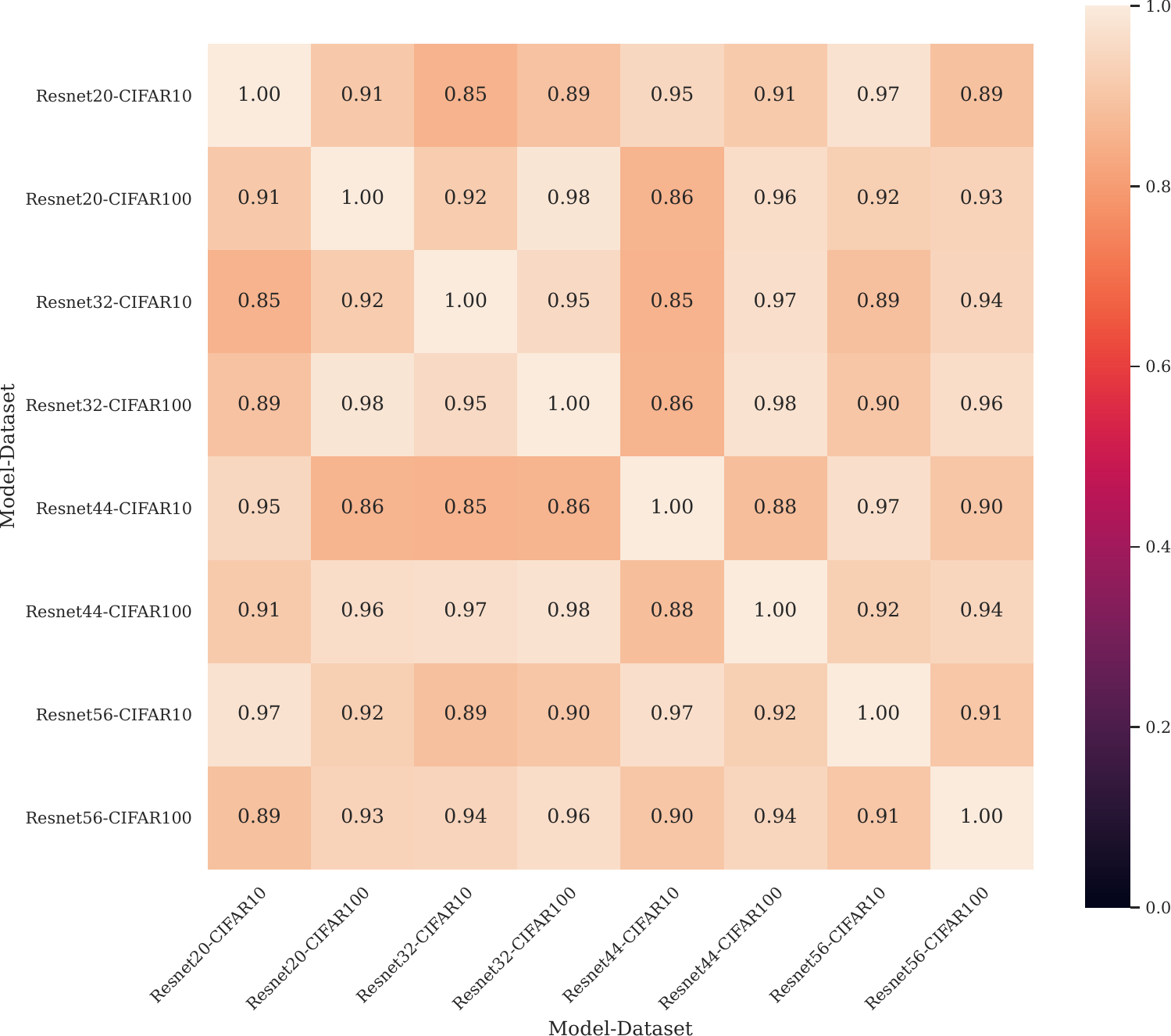} }}%

 \caption{Different ResNets trained on different datasets all learn highly consistent energy profiles in their first layer.}
  \label{fig:appendix-resnets}
\end{figure*}
\begin{figure*}[h]
  \centering
  \subfloat[\centering  Energy Profiles]{{\includegraphics[width=0.52\linewidth]{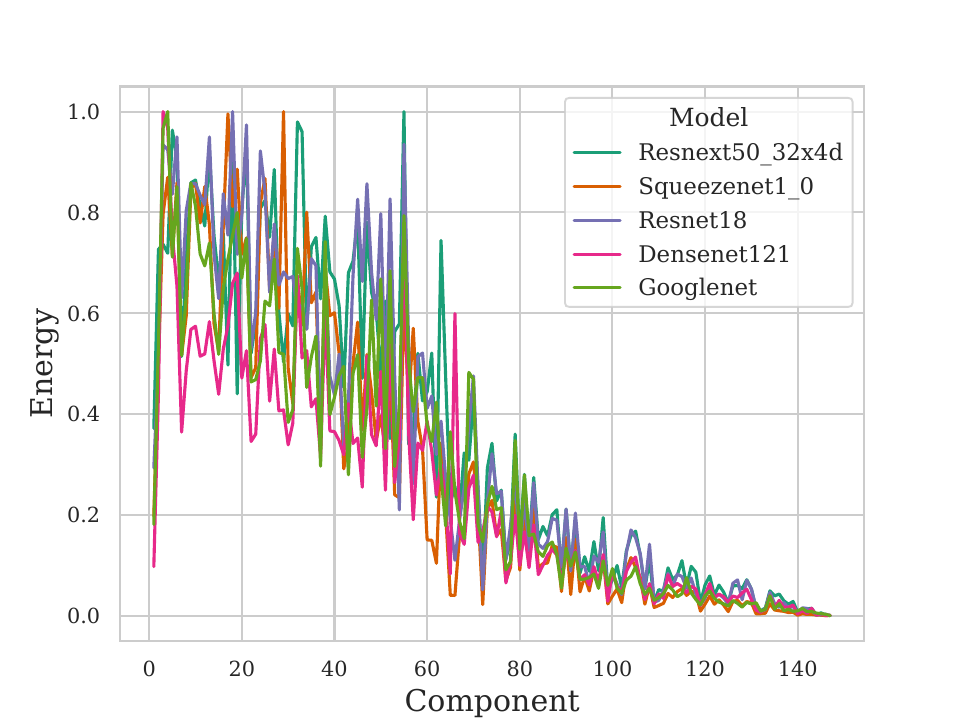} }}%
  \hfill
  \subfloat[\centering  Correlation Coefficients]{{\includegraphics[width=0.40\linewidth]{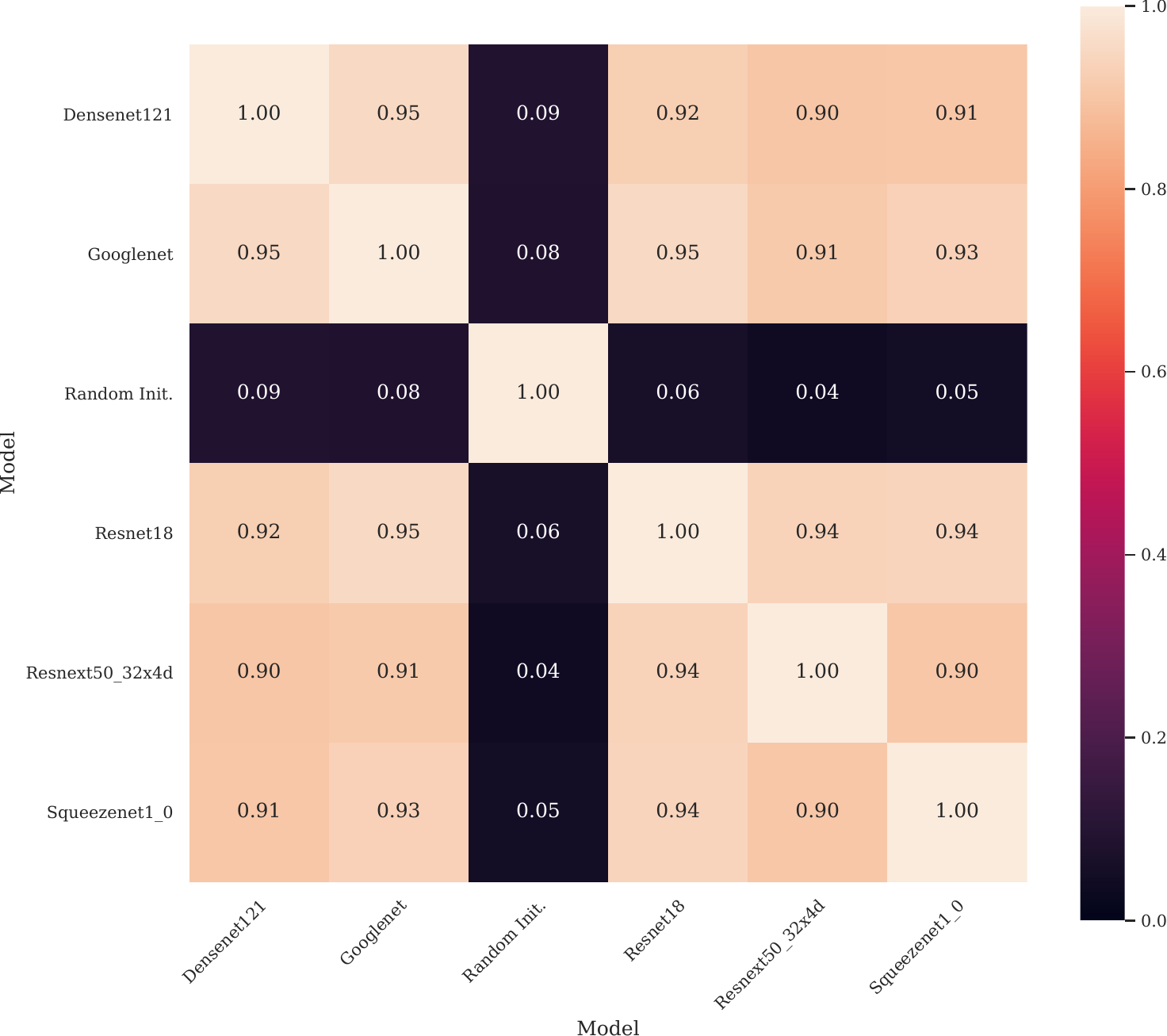} }}%

 \caption{Energy profiles of different models trained on ImageNet with filter size of $3 \times 7\times 7$.  Models are highly consistent yet differ from initialization. }
  \label{fig:appendix-imagenets}
\end{figure*}
\begin{figure*}[h]
  \centering
  \subfloat[\centering  Energy Profiles]{{\includegraphics[width=0.52\linewidth]{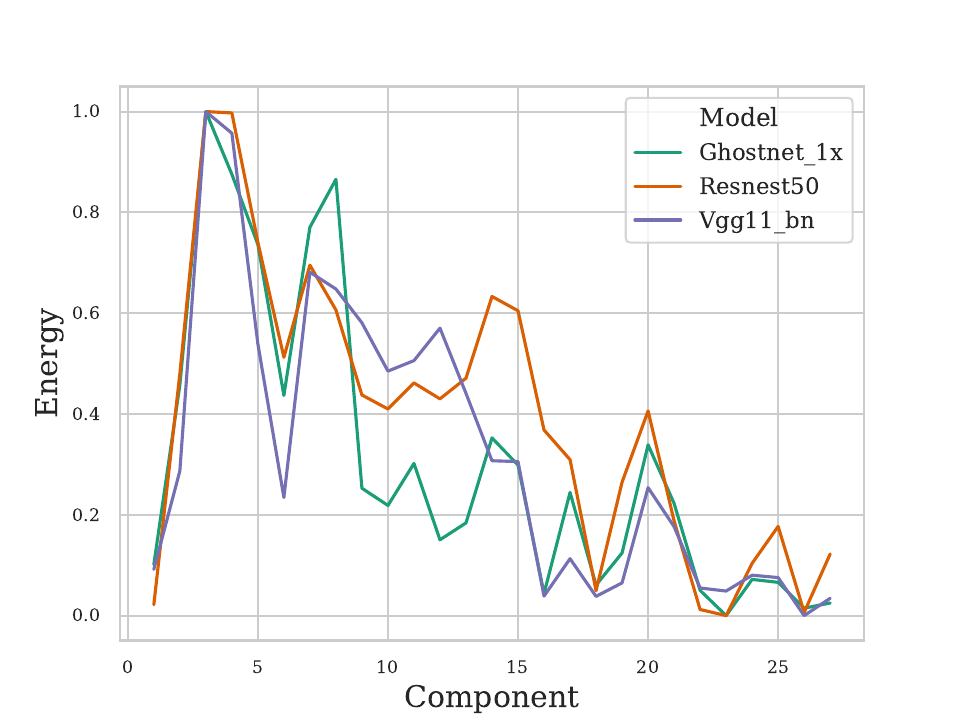} }}%
  \hfill
  \subfloat[\centering  Correlation Coefficients]{{\includegraphics[width=0.40\linewidth]{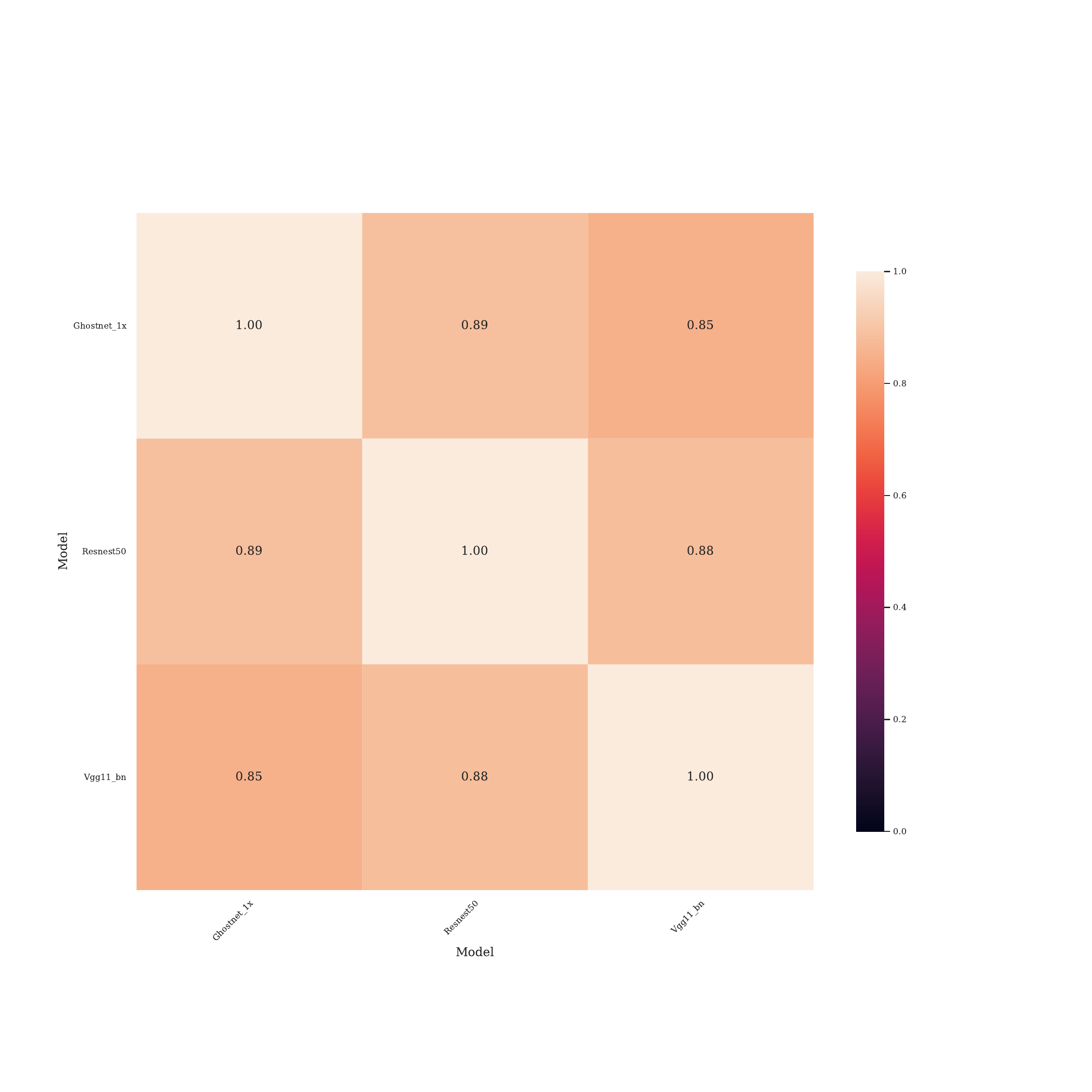} }}%

 \caption{Energy profiles of different models trained on ImageNet with filter size of $3 \times 3\times 3$. }
  \label{fig:appendix-imagenets3}
\end{figure*}

\begin{figure*}[h]
  \centering
  \subfloat[\centering  Energy Profiles]{{\includegraphics[width=0.52\linewidth]{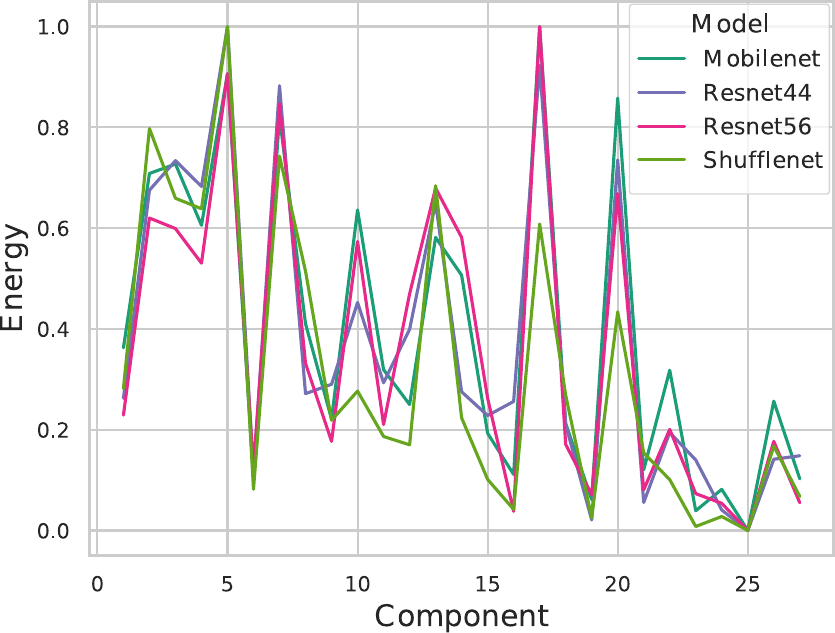} }}%
  \hfill
  \subfloat[\centering  Correlation Coefficients]{{\includegraphics[width=0.40\linewidth]{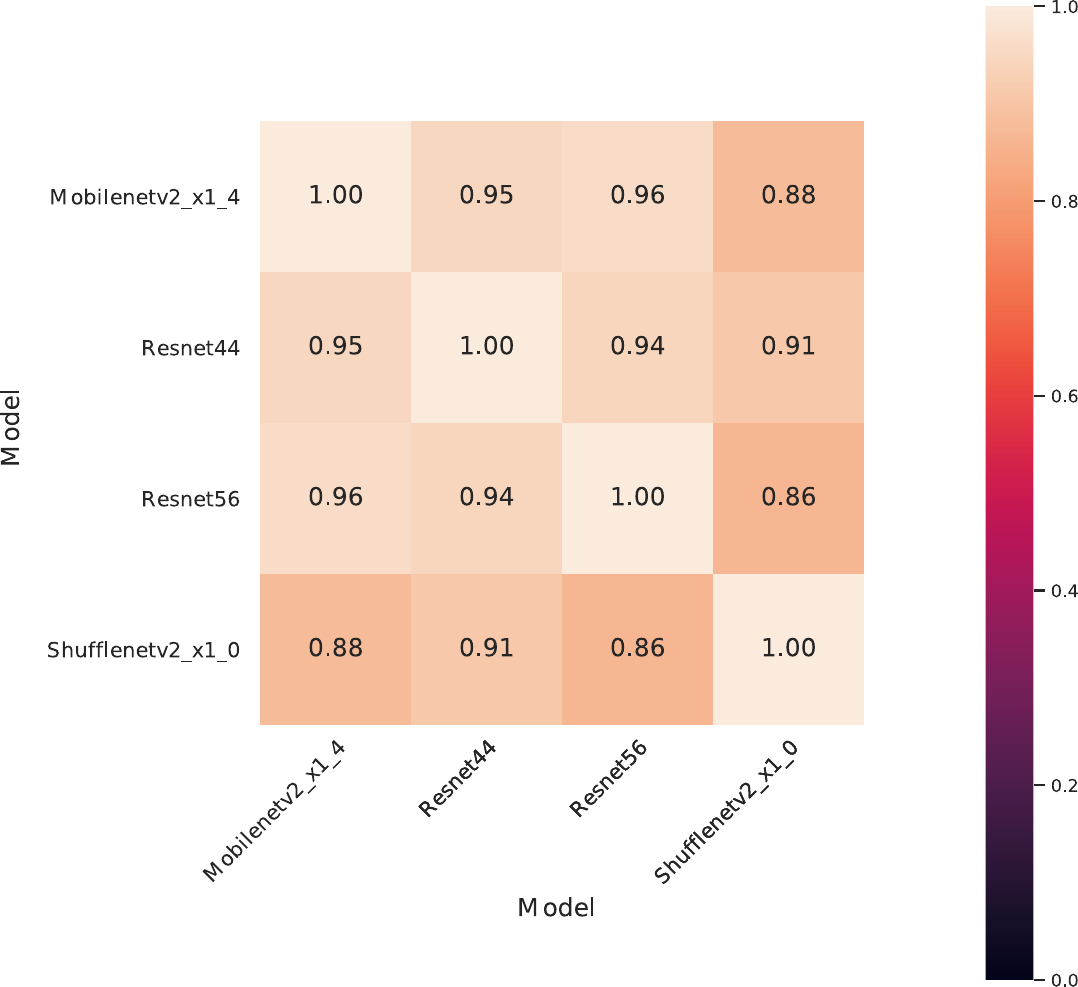} }}%

 \caption{Energy profiles of different models trained on CIFAR100 with filter size of $3 \times 3\times 3$. }
  \label{fig:appendix-cifar100}
\end{figure*}

\begin{figure*}[h]
  \centering
  \subfloat[\centering  Energy Profiles]{{\includegraphics[width=0.52\linewidth]{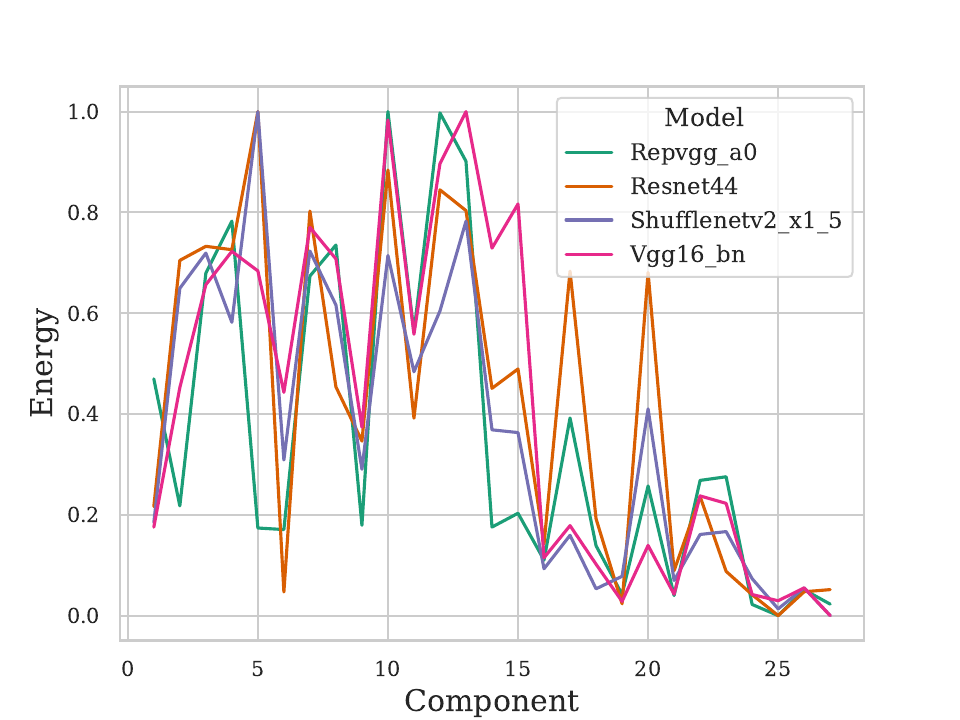} }}%
  \hfill
  \subfloat[\centering  Correlation Coefficients]{{\includegraphics[width=0.40\linewidth]{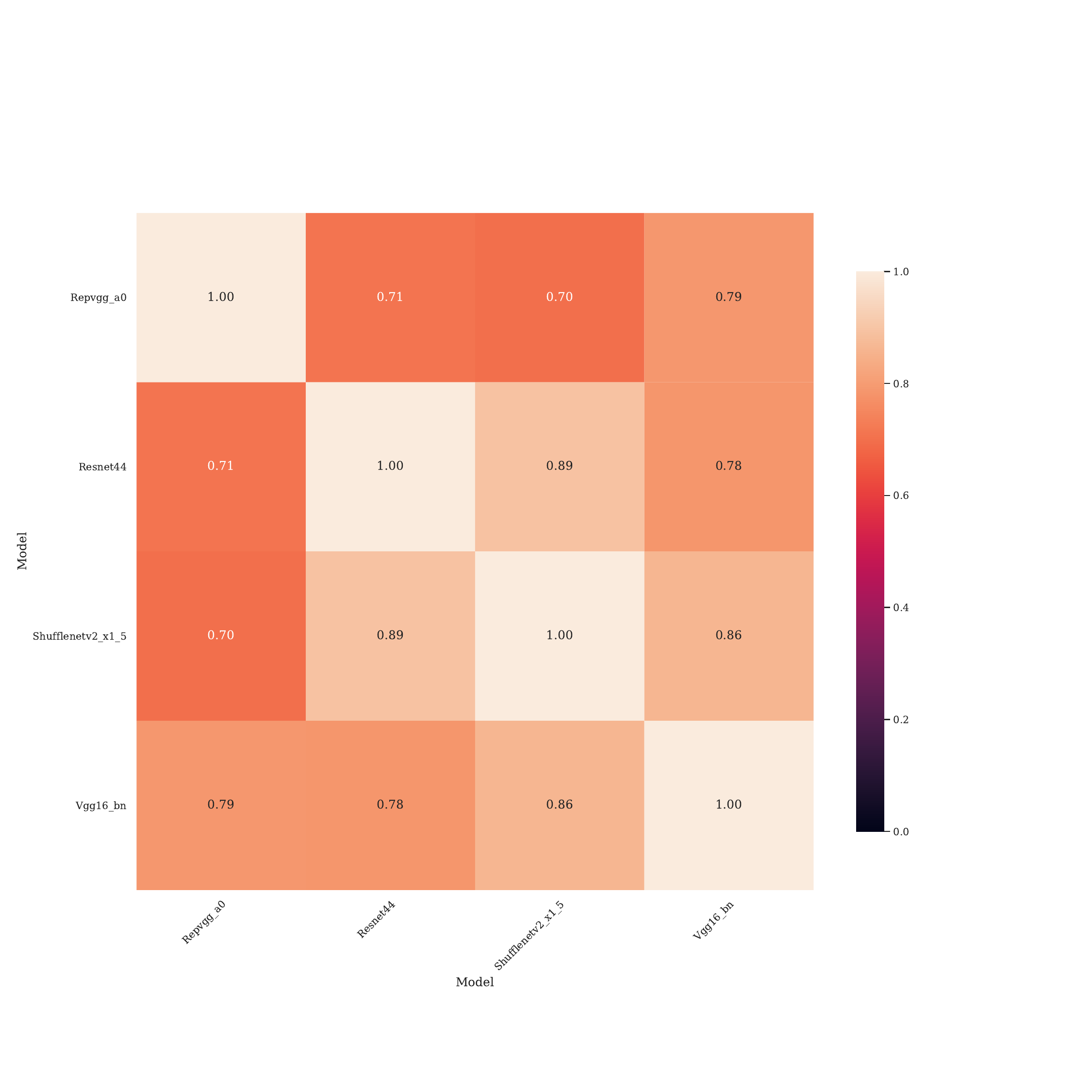} }}%

 \caption{Energy profiles of different models trained on CIFAR10 with filter size of $3 \times 3\times 3$. }
  \label{fig:appendix-cifar10}
\end{figure*}
\FloatBarrier
\section{Fitting Formula to Different Models}\label{appendix:formulafitexpanded}
To expand on the results in \cref{section:formulafit}, presented are more fits of the formula in \cref{to-provew-eq} to different models on different datasets. \cref{fig:appendix-imagenet-fit} depicts models downloaded from the PyTorch model hub, which are highly correlated with the theoretical formula. Meanwhile, a random initialization can hardly be explained using it.
\begin{figure*}[h]
  \centering
  \subfloat[\centering  DenseNet (0.9)]{{\includegraphics[width=0.32\linewidth]{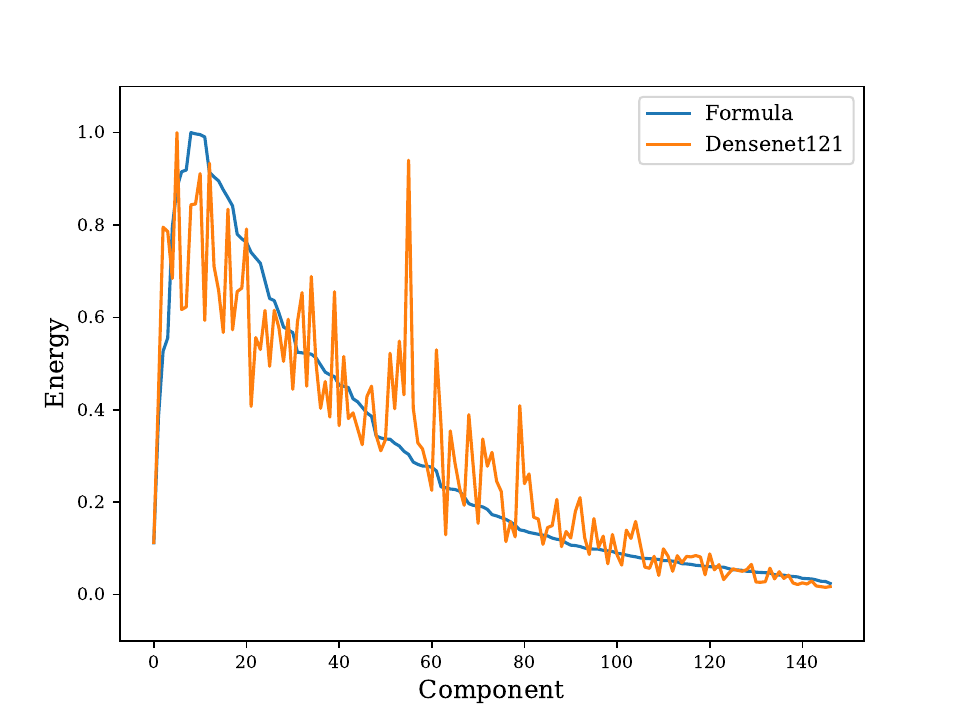} }}%
  \hfill
  \subfloat[\centering  GoogLeNet (0.9)]{{\includegraphics[width=0.32\linewidth]{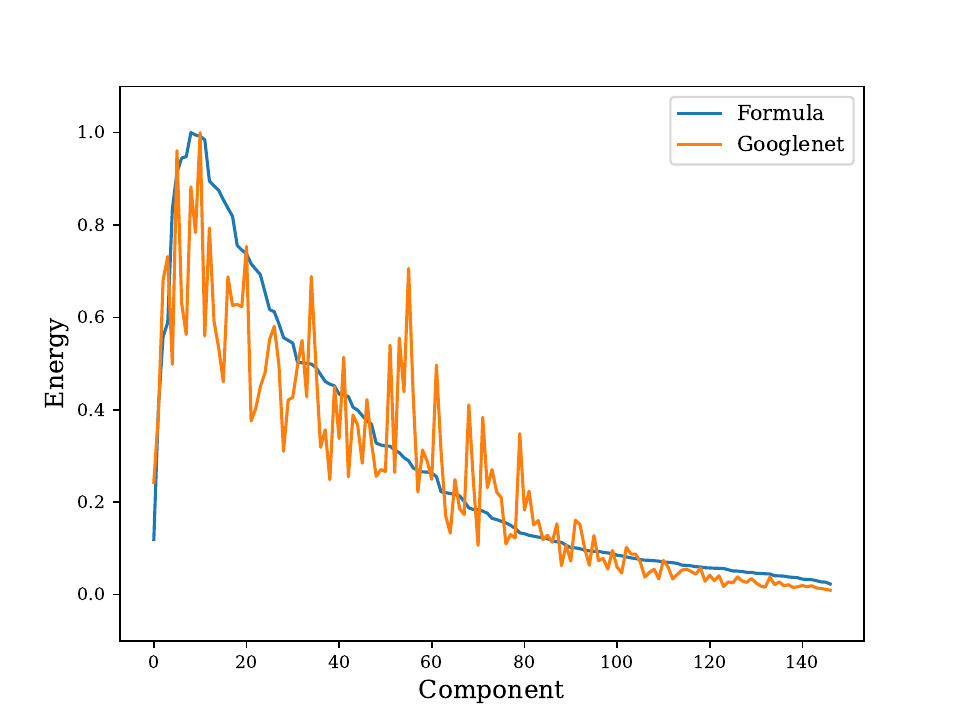} }}%
  \hfill
  \subfloat[\centering  ResNet18 (0.92)]{{\includegraphics[width=0.32\linewidth]{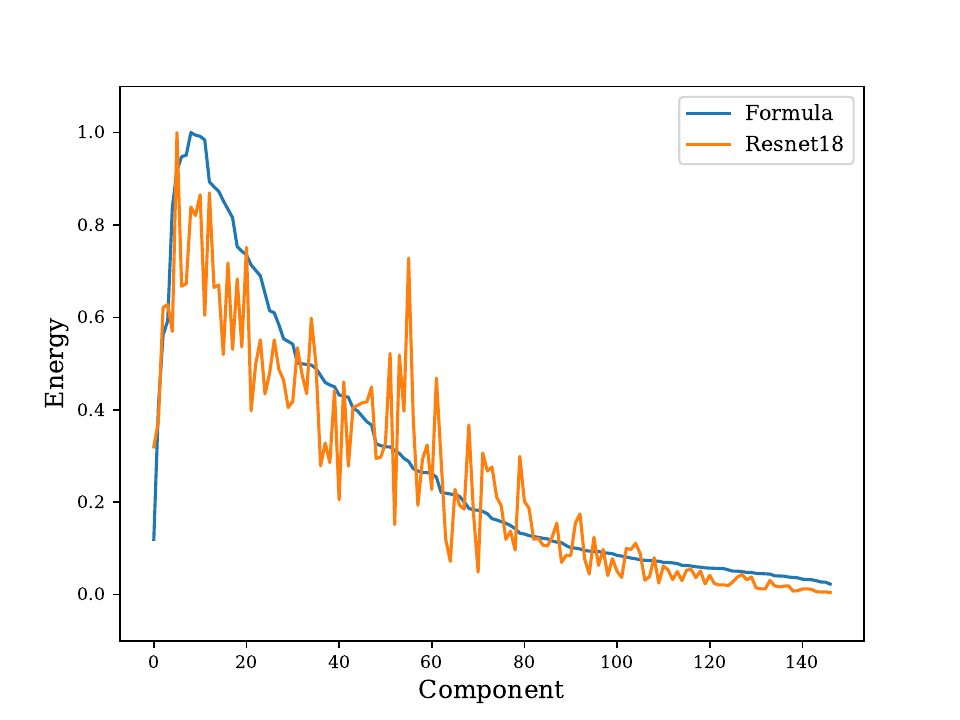} }}%
  \hfill
  \subfloat[\centering  ResNeXt (0.9)]{{\includegraphics[width=0.32\linewidth]{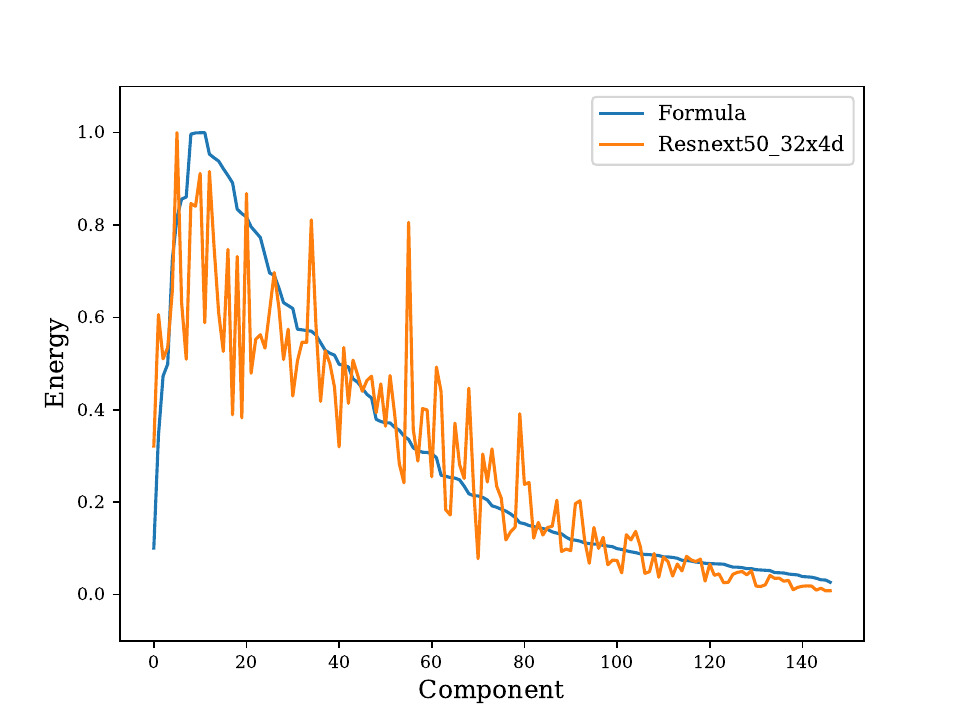} }}%
  \hfill
  \subfloat[\centering  SqueezeNet (0.9)]{{\includegraphics[width=0.32\linewidth]{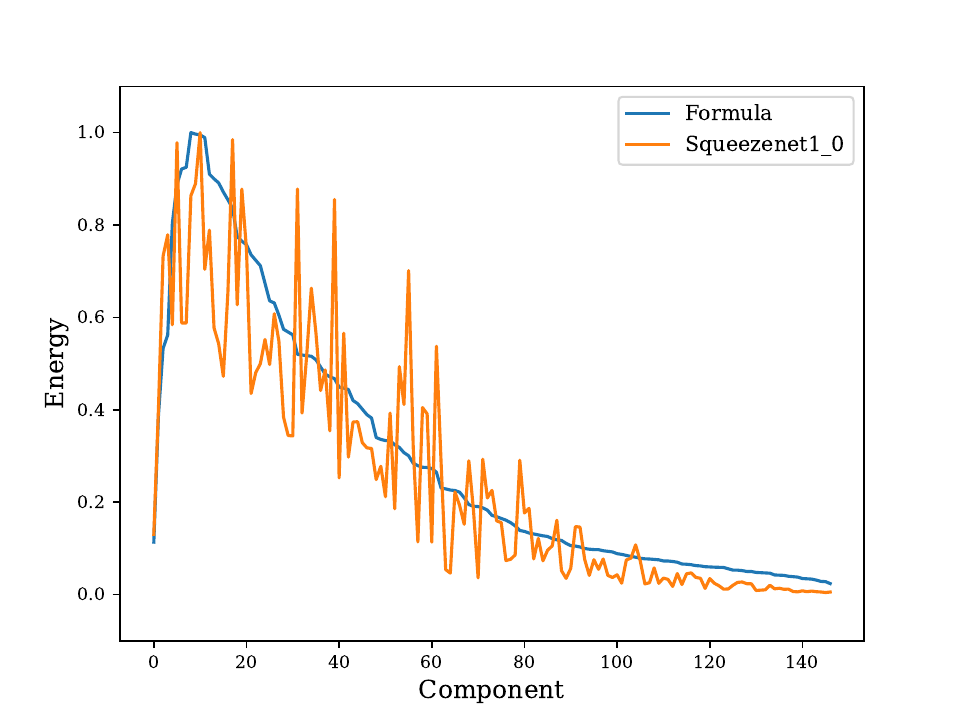} }}%
  \hfill
  \subfloat[\centering  Random Init. (0.11)]{{\includegraphics[width=0.32\linewidth]{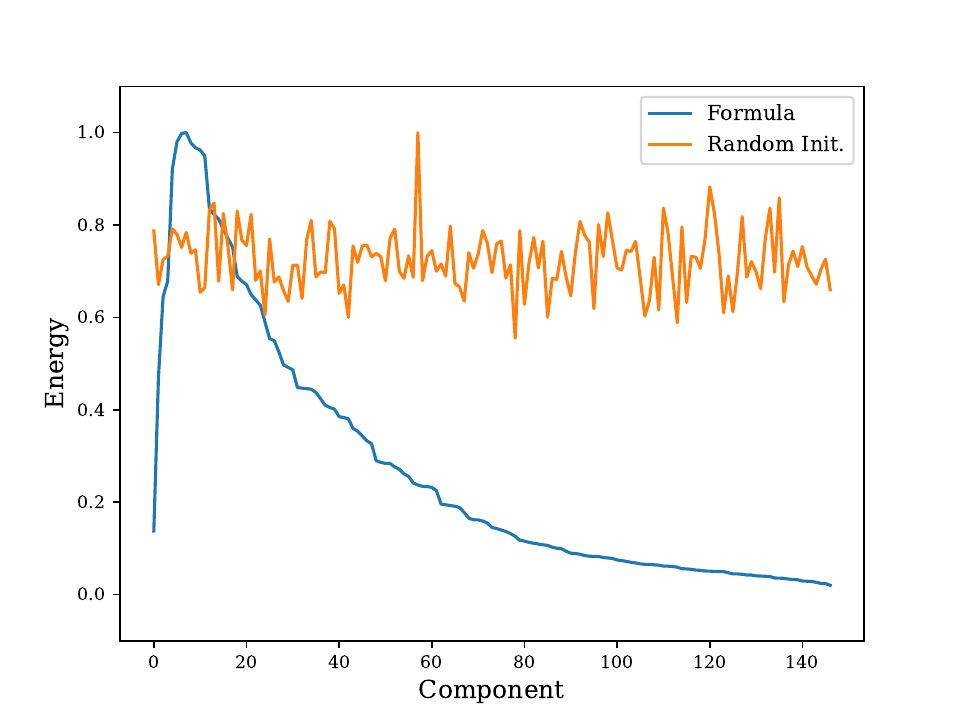} }}%

  \caption{Fitting \cref{to-provew-eq} to different models trained on ImageNet by searching over iterations. An example of a random initialization is attached for reference. Correlation coefficients in parentheses.}
  \label{fig:appendix-imagenet-fit}
\end{figure*}

\begin{figure*}[h]
  \centering
  \subfloat{{\includegraphics[width=0.32\linewidth]{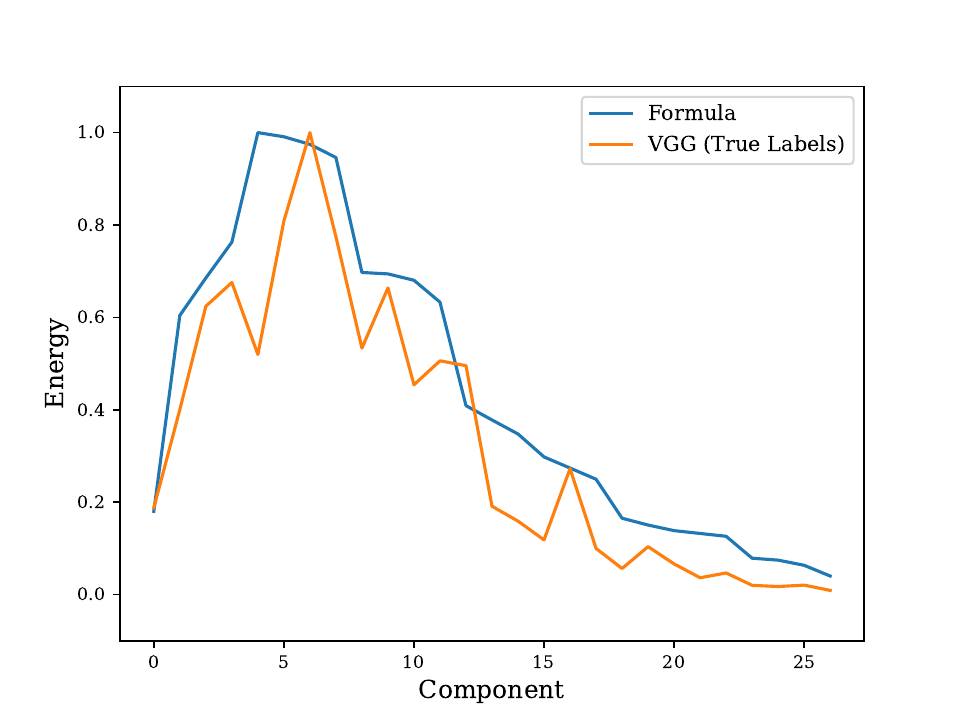} }}%
  \hfill
  \subfloat{{\includegraphics[width=0.32\linewidth]{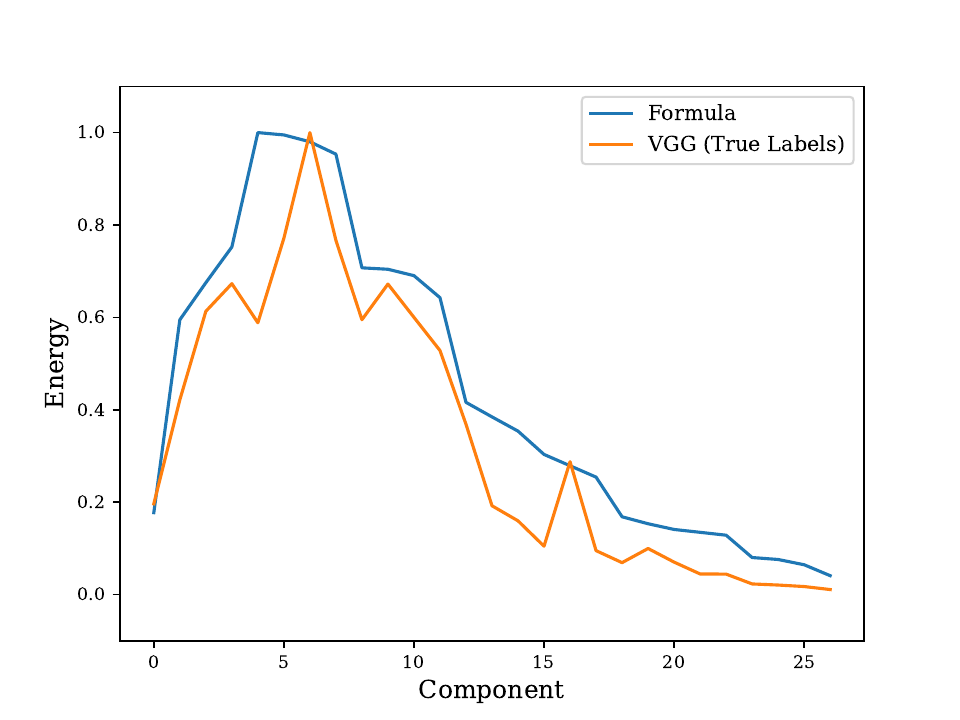} }}%
  \subfloat{{\includegraphics[width=0.32\linewidth]{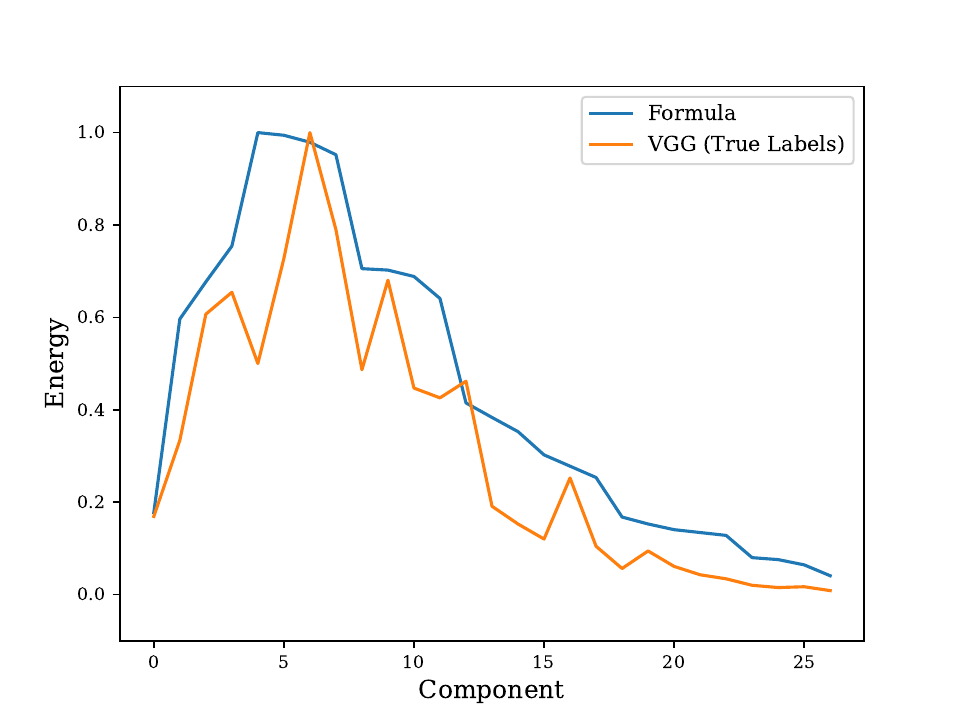} }}%

  \caption{More examples of fitting \cref{to-provew-eq} to VGG11 trained on CIFAR10 with different random seeds. Correlations are above 0.94.}
  \label{fig:appendix-imagenet-fit}
\end{figure*}

\begin{figure*}[h]
  \centering

  \subfloat{{\includegraphics[width=0.32\linewidth]{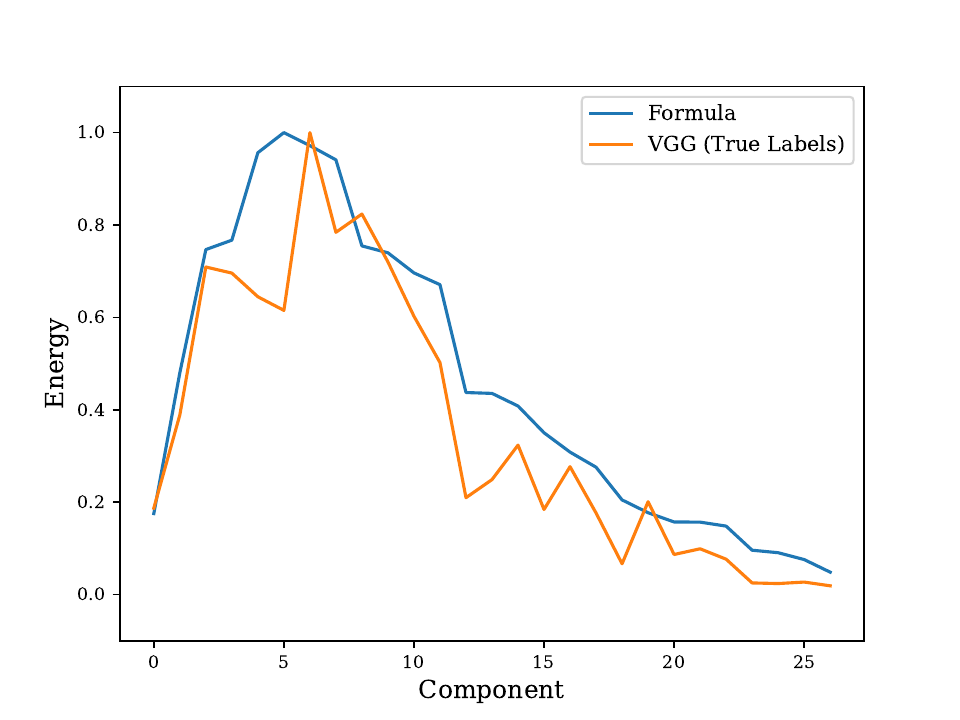} }}%
  \hfill
  \subfloat{{\includegraphics[width=0.32\linewidth]{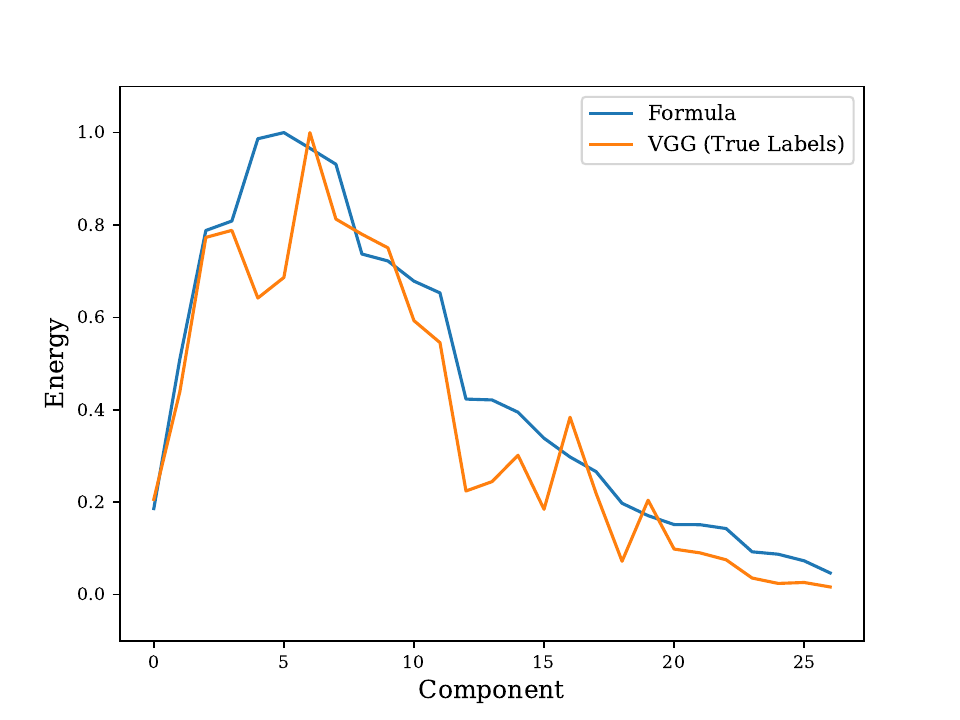} }}%
  \subfloat{{\includegraphics[width=0.32\linewidth]{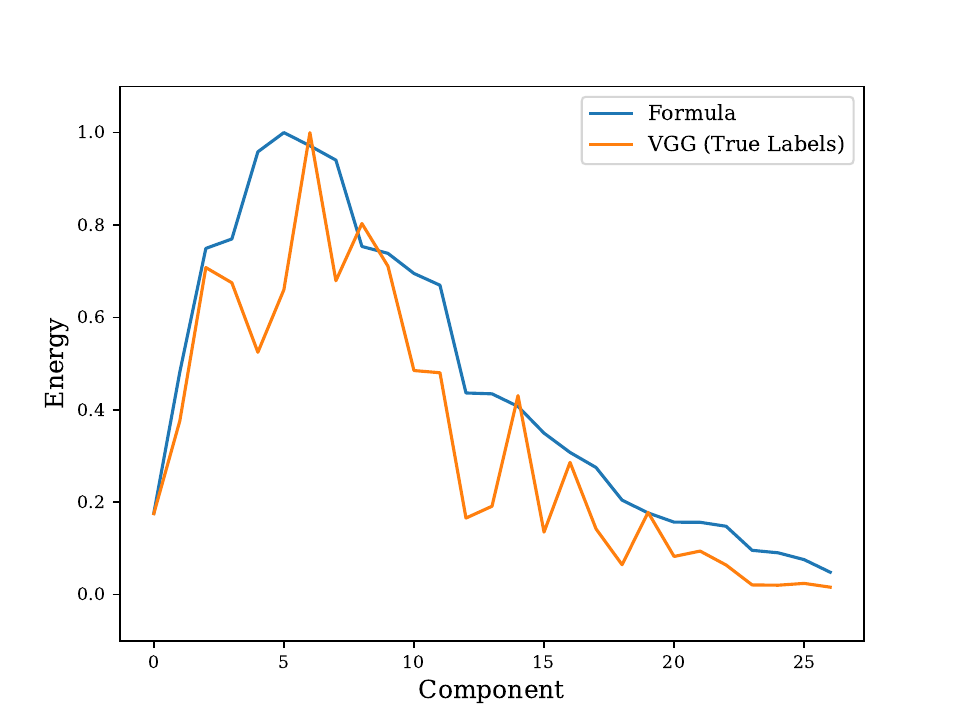} }}%
  \caption{Examples of fitting \cref{to-provew-eq} to VGG11 trained on CIFAR100 with different random seeds. Correlations are above 0.93.}
  \label{fig:appendix-imagenet-fit}
\end{figure*}

\FloatBarrier
\section{Effects of Changing Label and Image Statistics}\label{changingstats_appendix}
As explained in \cref{section:formulafit}, we conducted two experiments changing the input-output statistics and testing the effects on the learned energy profiles. According to \cref{theorem:maintheorem}, as long as the PCA components remain uncorrelated we expect models trained with true and random labels to remain consistent with each other and with the formula in \cref{to-provew-eq}.

\subsection{Introducing Correlation between Patches and Labels}\label{appendix:labelpatchcor}
\begin{figure*}[h]
  \centering
  \subfloat[\centering  Component 2 (0.44)]{{\includegraphics[width=0.45\linewidth]{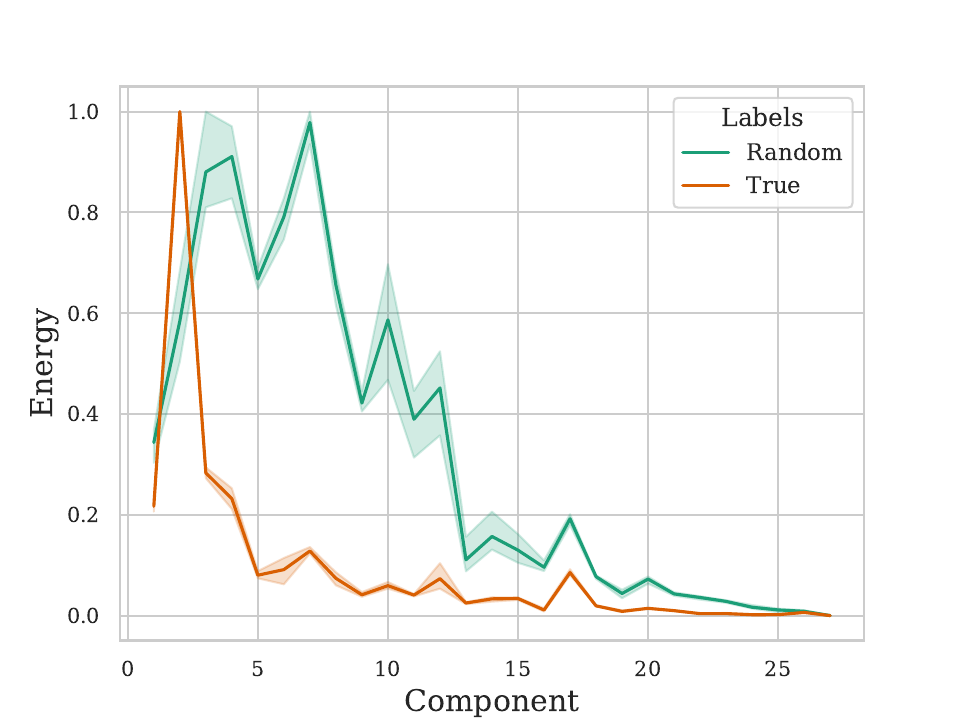} }}%
  \hfill
  \subfloat[\centering Component 3 (0.76)]{{\includegraphics[width=0.45\linewidth]{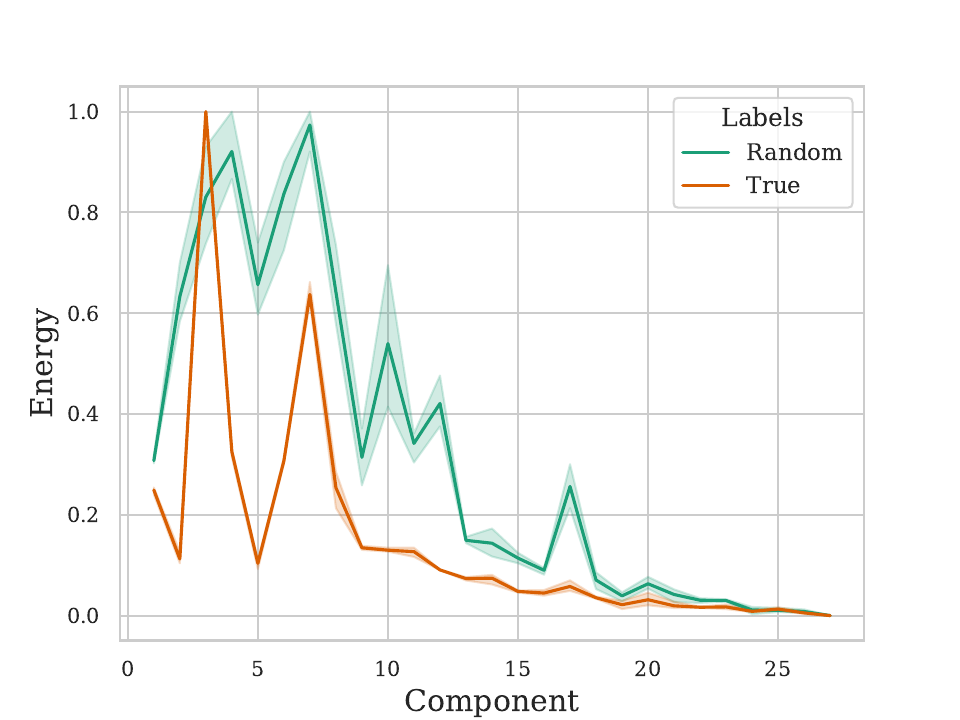}}}%
  \hfill
  \subfloat[\centering Component 16 (-0.06)]{{\includegraphics[width=0.45\linewidth]{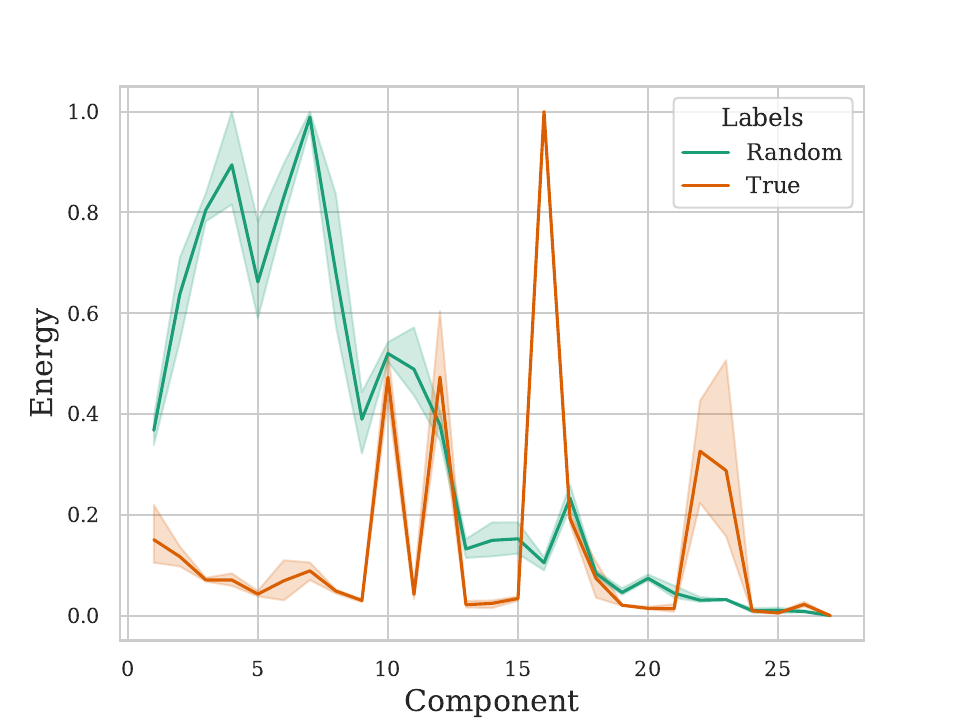}}}%
  \hfill
  \subfloat[\centering Component 26 (-0.24)]{{\includegraphics[width=0.45\linewidth]{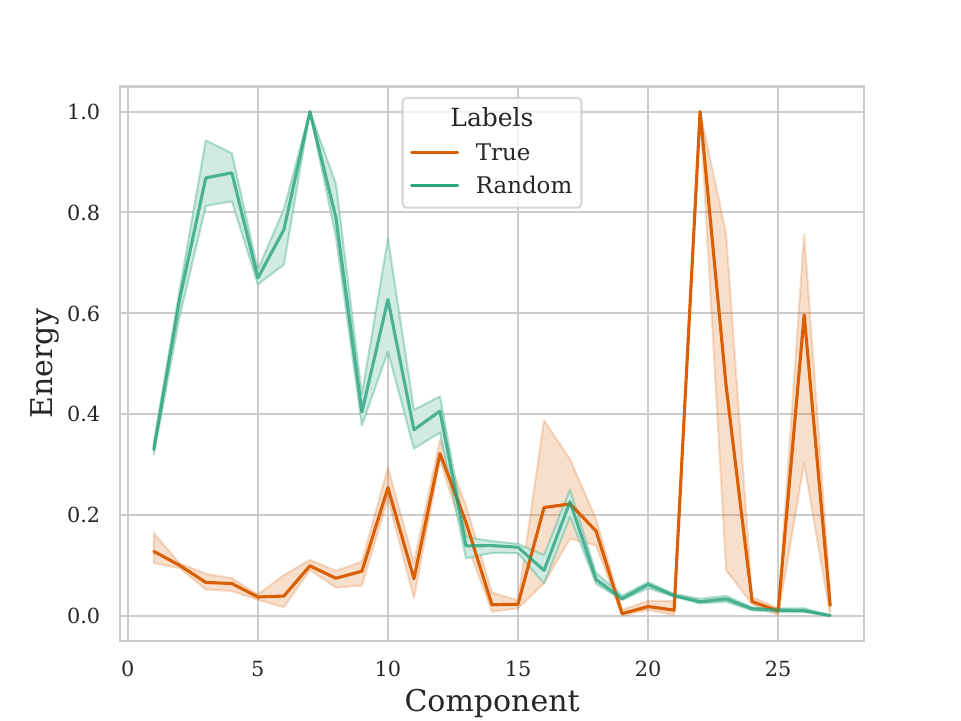}}}%

  \caption{ Training with true and random labels, when the true labels correspond to the image patch energy in different components (mean correlation in parenthesis). Once introducing correlation between patches and labels, profiles of true and random labels cease to correlate.}%
  \label{expandedlabelstatfig}
\end{figure*}

In the first, each image was labeled according to the energy w.r.t. a PCA component $u$. For an image $X$ with patches $P_1(X)...P_k(X)$ we calculated the quantity $\sum_{i=1}^k(P_i(X)^Tu)^2$ to be the total patch energy in direction $u$, and labeled $X$ according to the percentile of its energy (top 10\% of images w.r.t. their energy were labeled $y=1$, bottom 10\% were labeled $y=10$ and so on). \\
\cref{expandedlabelstatfig} shows that for different components, the correlation drops between the profiles of models trained with the new true labels and random labels. Notice, the decrease is more is larger when the labels are determined by components which aren't learned by the model with random labels. 

\FloatBarrier
\subsection{Changing the Patch Statistics}\label{appendix:eigentransform}
In this experiment, we changed the patch distribution. Let $u_1...u_d$ be the PCA components. Therefore each patch $P_i(X)$ of and image $X$ can be spanned as 
\begin{equation}
    P_i(X) = \sum_{j=1}^d \inprod{P_i(X), u_j}u_j
\end{equation}
We adjust the distribution by constant $\alpha > 1$ w.r.t. component $u_t$ by transforming:

\begin{equation}
     \sum_{j=1}^d \inprod{P_i(X), u_j}u_j \rightarrow  \alpha \inprod{P_i(X), u_t}u_t + \sum_{j\neq t} \inprod{P_i(X), u_j}u_j
\end{equation}
Therefore changing the patch PCA eigenvalue corresponding to $u_t$. We do this to all overlapping patches in the dataset (therefore with no affect to the correlation between patches and labels), and adjust the stride in the first layer to avoid issues with the overlap. \\

\begin{figure*}[h]
  \centering
    \subfloat[\centering  Component 0 (True)]{{\includegraphics[width=0.45\linewidth]{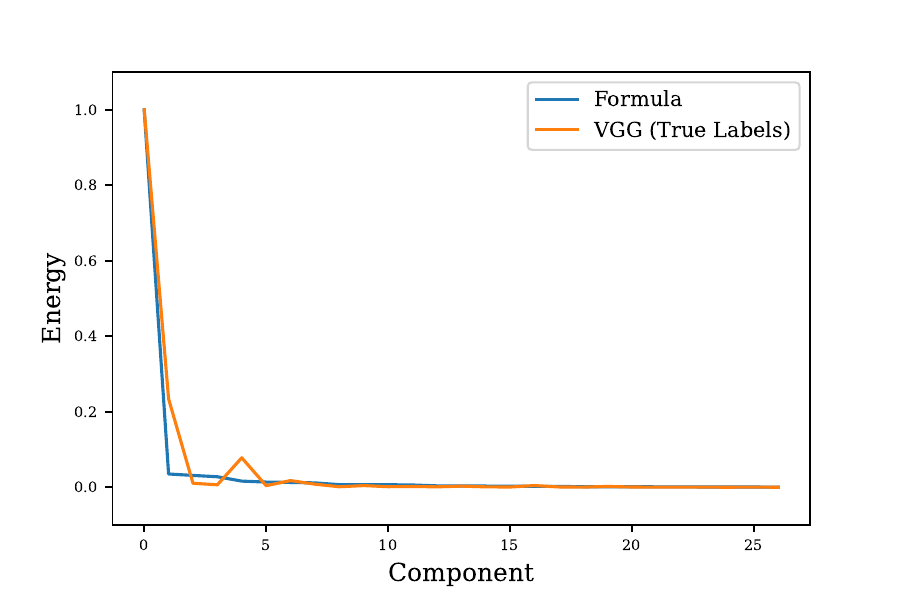} }}%
  \hfill
  \subfloat[\centering Component 0 (Random)]{{\includegraphics[width=0.45\linewidth]{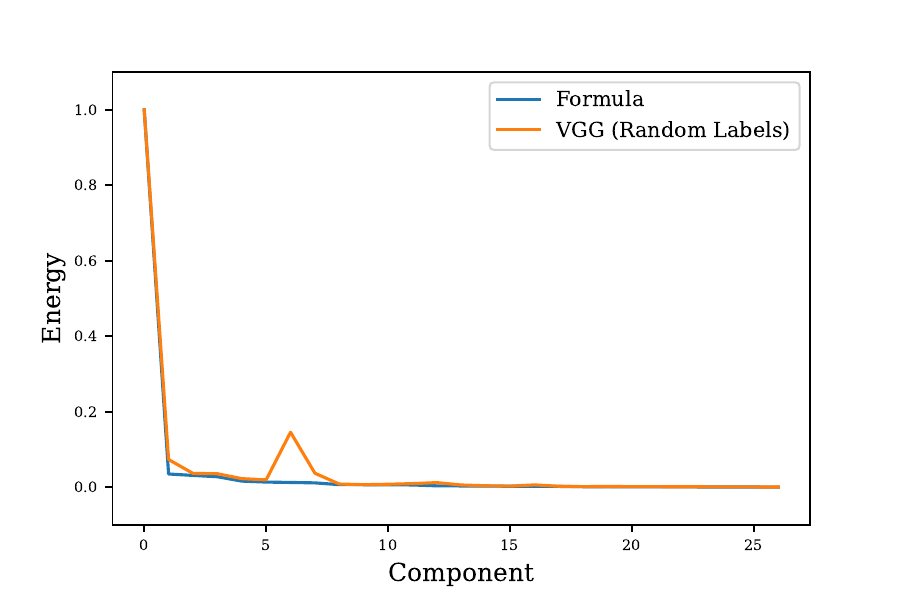}}}%
  \hfill
  \subfloat[\centering  Component 15 (True)]{{\includegraphics[width=0.45\linewidth]{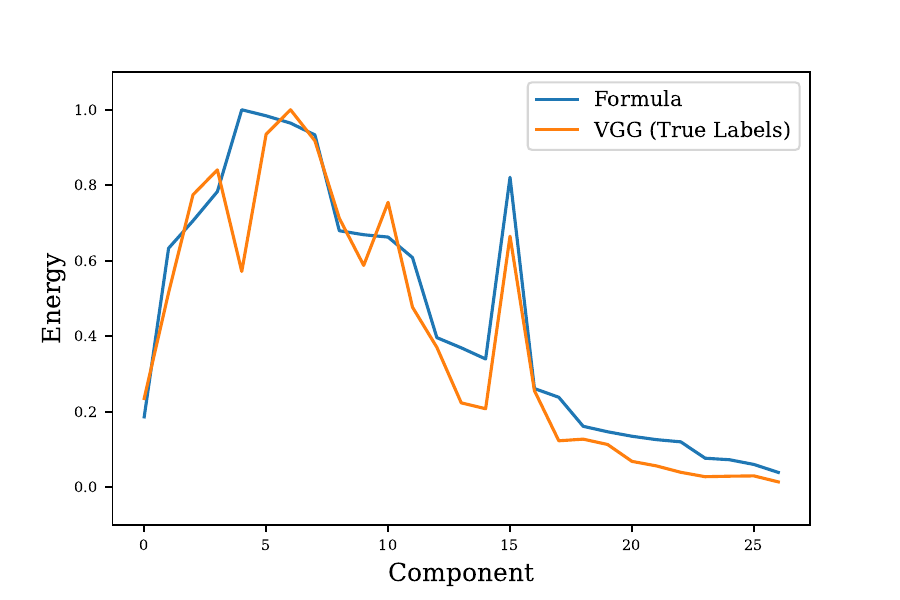} }}%
  \hfill
  \subfloat[\centering Component 15 (Random)]{{\includegraphics[width=0.45\linewidth]{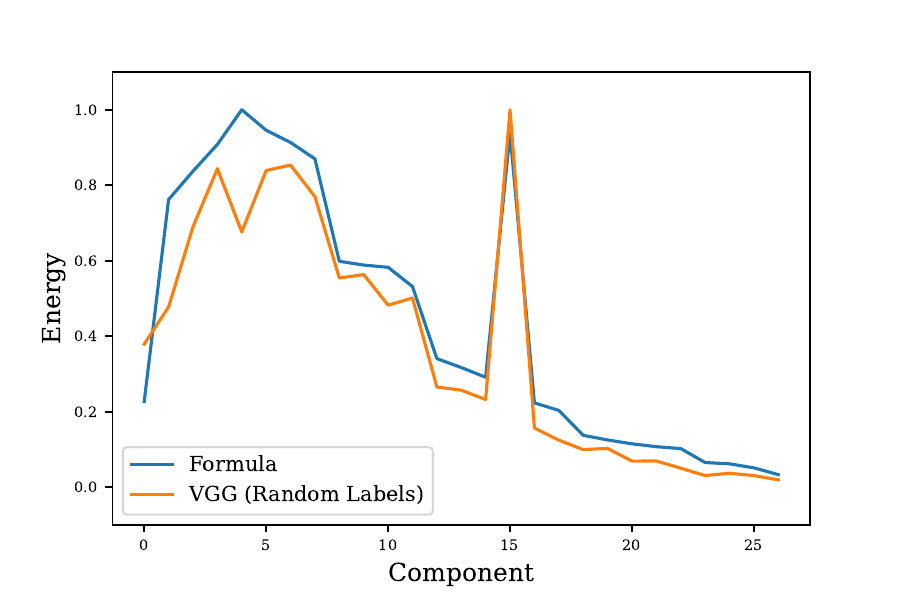}}}%
  \hfill
  \subfloat[\centering Component 25 (True)]{{\includegraphics[width=0.45\linewidth]{figures/correlation_changes/patch_energy_change/label_1_ind_25_fac_10_cor94.pdf}}}%
  \hfill
  \subfloat[\centering Component 25 (True)]{{\includegraphics[width=0.45\linewidth]{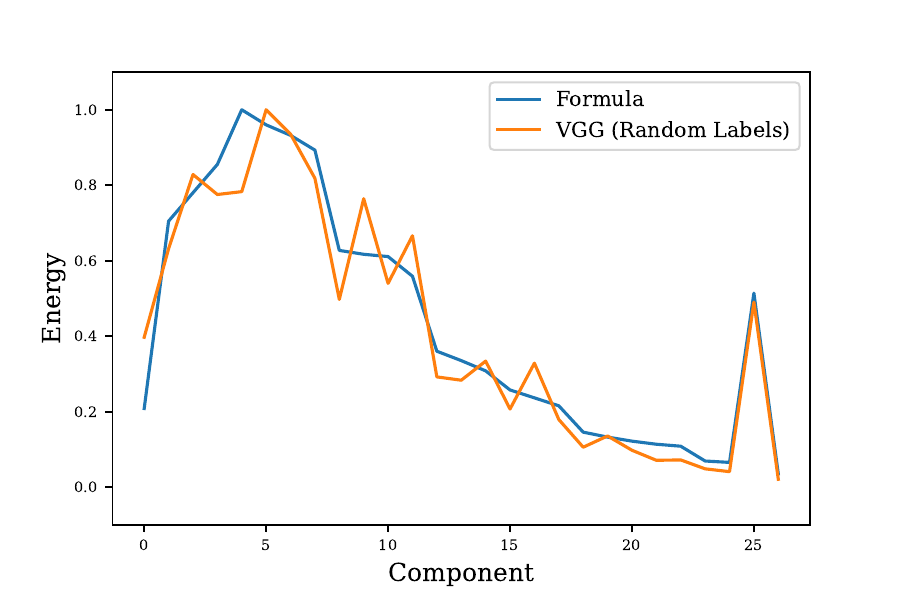}}}%

  \caption{ Changing the eigenvalues corresponding to different PCA component for true and random labels. The profiles for both sets of labels are highly similar and can be explained by the analytic formula.}%
  \label{changepatchstats}
\end{figure*}

\cref{changepatchstats} shows that indeed true and random labels after applying the transformation are still similar, and both can be explained by applying the same transformation to the eigenvalues used to calculate the analytic formula. 

\FloatBarrier
\section{Visual Similarity of Filters in the First Layer}\label{appendix:visual-similarity}

\begin{figure*}[h]
  \centering
  \subfloat[\centering Initialized ResNet18 Filters ]{{\label{initfilters}\includegraphics[width=0.45\linewidth]{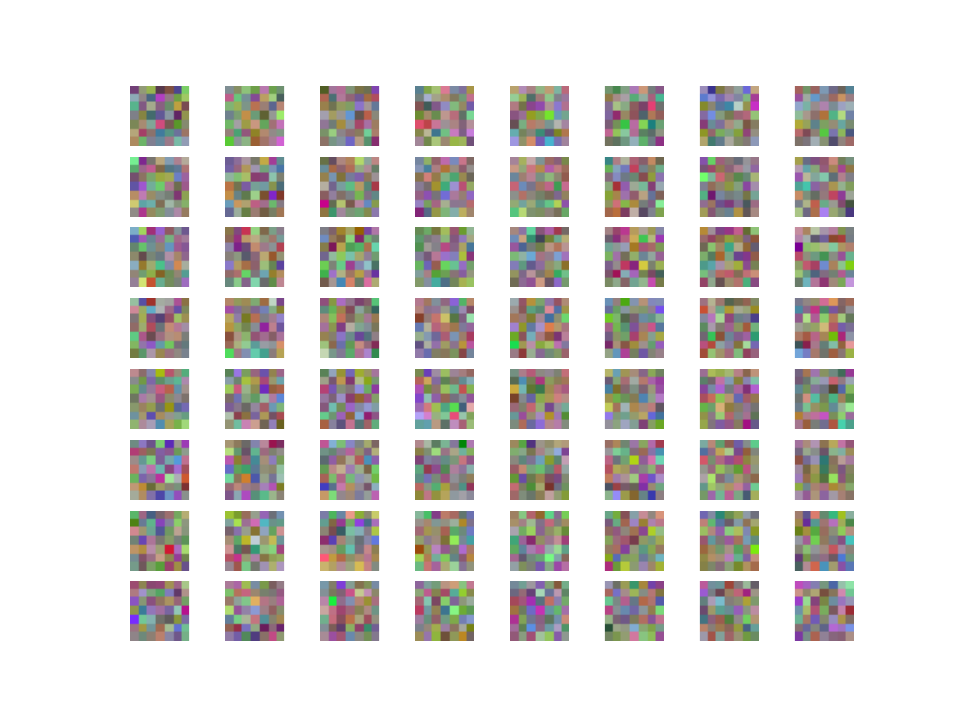} }}%
  \hfill
  \subfloat[\centering Trained ResNet18 Filters]{{\label{resfilters}\includegraphics[width=0.45\linewidth]{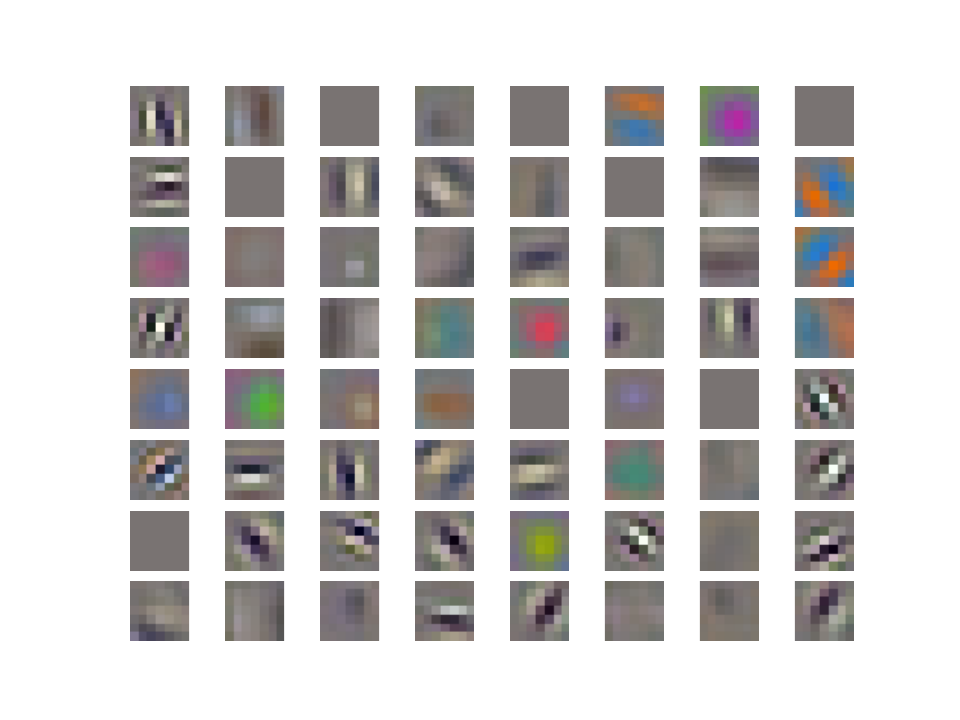}}}%
  \hfill
  \subfloat[\centering Trained GoogleNet Filters]{{\label{googfilters}\includegraphics[width=0.45\linewidth]{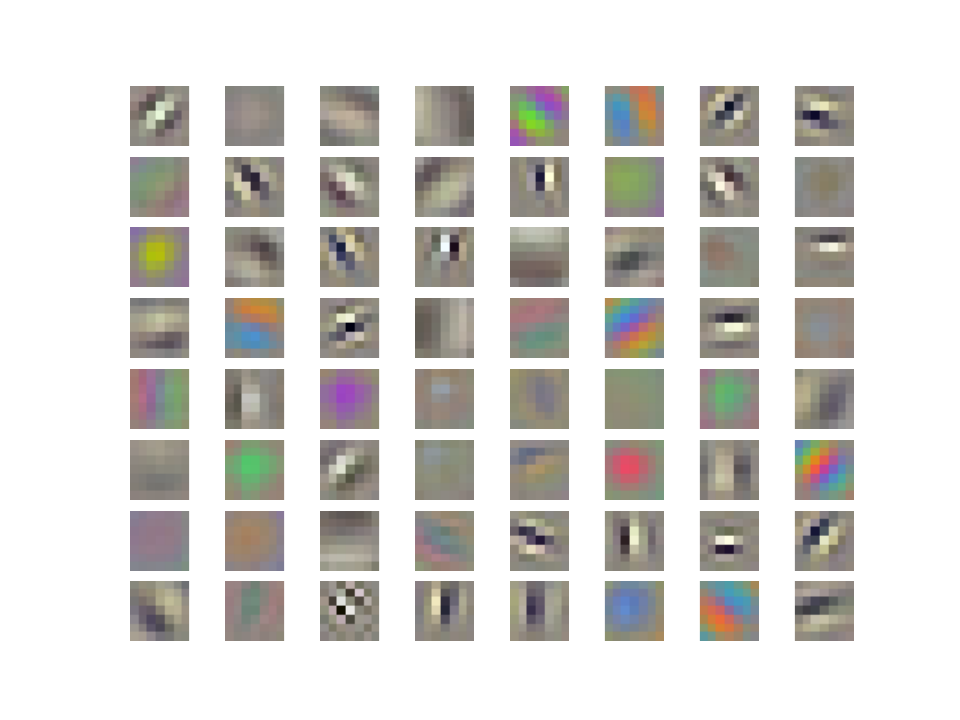}}}%
  \hfill
  \subfloat[\centering Trained DenseNet Filters]{{\label{densefilters}\includegraphics[width=0.45\linewidth]{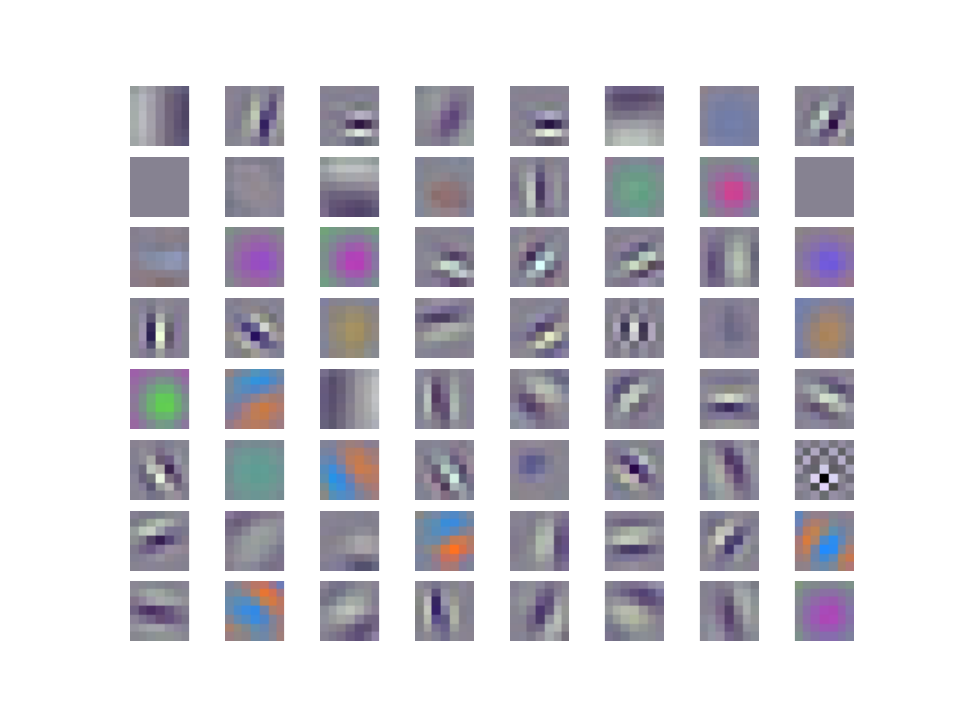}}}%

  \caption{ Different CNNs (\ref{resfilters}, \ref{googfilters}) trained on ImageNet learn a highly consistent first layer despite using different architectures. These filters are very different from the initial, random filters showing that consistent representation learning has occurred. In this paper we quantify this consistency and seek to understand its source. }%
  \label{fig:vizsim}
\end{figure*}

As has already been pointed out by \cite{AlexNet,bipartitematch} filters learned by CNNs learn visually similar filters. For the readers convenience, \cref{fig:vizsim} displays filters taken from different networks trained on ImageNet. Notice these are noticably differ from a random initialization.

\FloatBarrier
\section{CNNs with Frozen First Layer}\label{frozen_appendix}
\begin{figure}[t]
\vskip 0.2in
   \centering
   \subfloat[\centering Train Loss ]{{\label{trainloss}\includegraphics[width=0.45\linewidth]{figures/frozen_vs_unfrozen_figures/Train_Loss.pdf} }}%
  \hfill   
   \subfloat[\centering Validation Loss]{{\label{valloss1}\includegraphics[width=0.45\linewidth]{figures/frozen_vs_unfrozen_figures/Validation_Loss.pdf}}}%
     \hfill   
   \subfloat[\centering Train Accuracy]{{\label{valloss2}\includegraphics[width=0.45\linewidth]{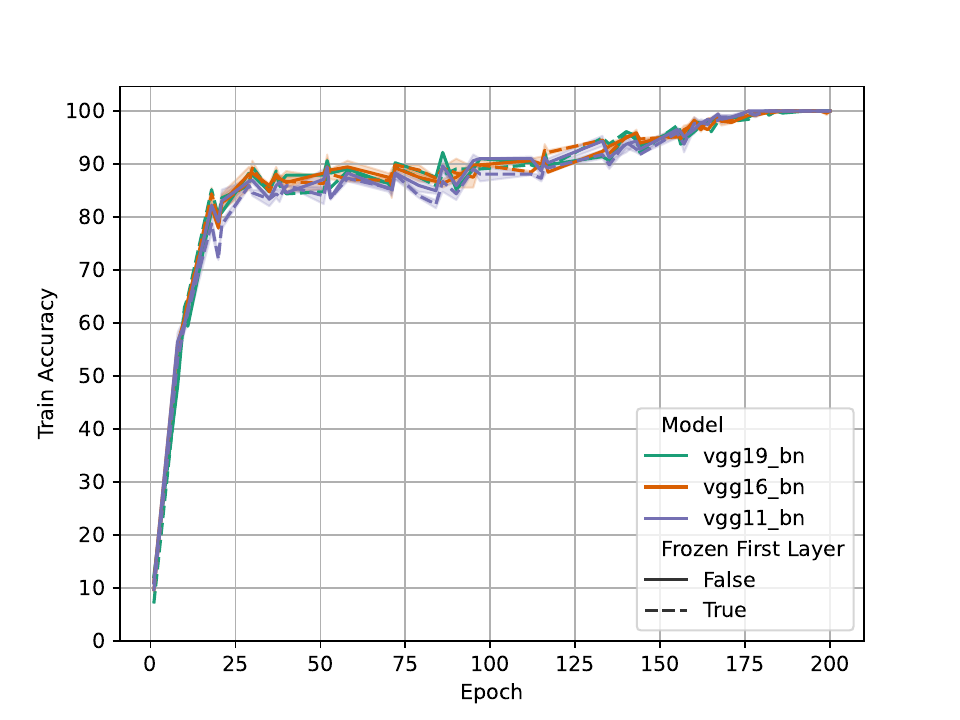}}}%
     \hfill   
   \subfloat[\centering Validation Accuracy]{{\label{valloss3}\includegraphics[width=0.40\linewidth]{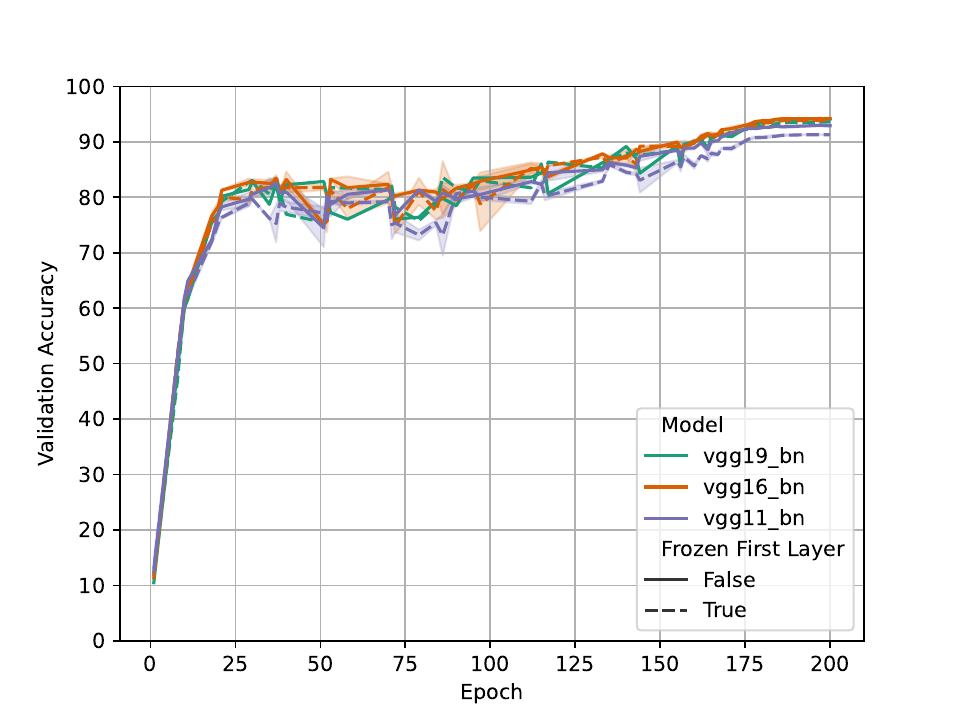}}}%
   \caption{Loss and accuracy metrics for VGGs of different depths, with and without a frozen layer, on CIFAR10, as function of iteration.}\label{frozend_appendix_fig}
      \vskip 0.1in
\end{figure}

To expand on the result on VGGs with a frozen first layer we attach here the full set of training results. As can be seen in \cref{frozend_appendix_fig}, as the depth of a network increases the difference between the model with a frozen first layer and a learnt one is almost indistinguishable - both in terms of accuracy and loss.

\end{document}